\documentclass[runningheads]{llncs}
\usepackage{graphicx}

\usepackage{tikz}
\usepackage{comment}
\usepackage{amsmath,amssymb} 
\usepackage{color}

\usepackage{algorithm}
\usepackage{algorithmic}
\newcommand{\tabincell}[2]{\begin{tabular}{@{}#1@{}}#2\end{tabular}}
\usepackage{verbatim}
\usepackage{multirow}
\usepackage[colorlinks,linkcolor=blue]{hyperref}

\begin{document}
\pagestyle{headings}
\mainmatter
\def\ECCVSubNumber{2308}  

\title{TF-NAS: Rethinking Three Search Freedoms of Latency-Constrained Differentiable Neural Architecture Search} 

\titlerunning{Rethinking Three Search Freedoms of Differentiable NAS}
%
\author{Yibo Hu\inst{1,2} \and
Xiang Wu\inst{1} \and
Ran He\inst{1}\thanks{corresponding author}}
\authorrunning{Y. Hu et al.}
%
\institute{CRIPAC \& NLPR, CASIA, Beijing, China \and
JD AI Research, Beijing, China \\
\email{\{huyibo871079699,alfredxiangwu\}@gmail.com}, \email{rhe@nlpr.ia.ac.cn}}
\maketitle

\begin{abstract}
With the flourish of differentiable neural architecture search (NAS), automatically searching latency-constrained architectures gives a new perspective to reduce human labor and expertise. However, the searched architectures are usually suboptimal in accuracy and may have large jitters around the target latency. In this paper, we rethink three freedoms of differentiable NAS, i.e. operation-level, depth-level and width-level, and propose a novel method, named Three-Freedom NAS (TF-NAS), to achieve both good classification accuracy and precise latency constraint. For the operation-level, we present a \textbf{bi-sampling} search algorithm to moderate the operation collapse. For the depth-level, we introduce a \textbf{sink-connecting} search space to ensure the mutual exclusion between skip and other candidate operations, as well as eliminate the architecture redundancy. For the width-level, we propose an \textbf{elasticity-scaling} strategy that achieves precise latency constraint in a progressively fine-grained manner. Experiments on ImageNet demonstrate the effectiveness of TF-NAS. Particularly, our searched TF-NAS-A obtains 76.9\% top-1 accuracy, achieving state-of-the-art results with less latency. 
Code is available at \href{https://github.com/AberHu/TF-NAS}{https://github.com/AberHu/TF-NAS}.
\end{abstract}

\section{Introduction}
With the rapid developments of deep learning, ConvNets have been the \textit{de facto} method for various computer vision tasks. It takes a long time and substantial effort to devise many useful models~\cite{DBLP:conf/cvpr/HeZRS16,DBLP:conf/cvpr/HuSS18,DBLP:conf/cvpr/HuangLMW17,DBLP:conf/eccv/MaZZS18,DBLP:conf/cvpr/SandlerHZZC18,DBLP:conf/cvpr/SzegedyVISW16}, boosting significant improvements in accuracy. However, instead of accuracy improvement, designing efficient ConvNets with specific resource constraints (e.g. FLOPs, latency, energy) is more important in practice. Manual design requires a huge number of exploratory experiments, which is time-consuming and labor intensive. Recently, Neural Architecture Search (NAS) has attracted lots of attentions~\cite{DBLP:conf/eccv/LiuZNSHLFYHM18,DBLP:conf/iclr/LiuSY19,DBLP:conf/aaai/RealAHL19,DBLP:conf/cvpr/WuDZWSWTVJK19,DBLP:conf/iclr/ZophL17}. It learns to automatically discover resource-constrained architectures, which can achieve better performance than hand-craft architectures.

Most NAS methods are based on reinforcement learning (RL)~\cite{DBLP:conf/cvpr/TanCPVSHL19,DBLP:conf/iclr/ZophL17,DBLP:conf/cvpr/ZophVSL18} or evolutionary algorithms (EA)~\cite{DBLP:journals/corr/abs-1908-06022,DBLP:conf/cvpr/DaiZWYSWDHWJVUJ19,DBLP:conf/aaai/RealAHL19}, leading to expensive or even unaffordable computing resources. Differentiable NAS~\cite{DBLP:conf/iclr/CaiZH19,DBLP:conf/iclr/LiuSY19,DBLP:conf/cvpr/WuDZWSWTVJK19} couples architecture sampling and training into a supernet to reduce huge resource overhead. This supernet supports the whole search space with three freedoms, including the operation-level, the depth-level and the width-level freedoms. 
However, due to the various combinations of search freedoms and the coarse-grained discreteness of search space, differentiable NAS often makes the searched architectures suboptimal with specific resource constraints. For example, setting the GPU latency constraint to 15ms and carefully tuning the trade-off parameters, we search for architectures based on the latency objective from ProxylessNAS~\cite{DBLP:conf/iclr/CaiZH19}. The searched architecture has 15.76ms GPU latency, exceeding the target by a large margin. More analyses are presented in Sec.~\ref{elasticity-scaling_exp}.

To address the above issue, in this paper, we first rethink the operation-level, the depth-level and the width-level search freedoms, tracing back to the source of search instability. For the operation-level, we observe operation collapse phenomenon, where the search procedure falls into some fixed operations. To alleviate such collapse, we propose a bi-sampling search algorithm. For the depth-level, we analyze the special role of skip operation and explain the mutual exclusion between skip and other operations. Furthermore, we also illustrate architecture redundancy by a simple case study in Fig.~\ref{fig:depth_illustration}. To address these phenomena, we design a sink-connecting search space for NAS. For the width-level, we explore that due to the coarse-grained discreteness of search space, it is hard to search target architectures with precise resource constraints (e.g. latency). Accordingly, we present an elasticity-scaling strategy that progressively refines the coarse-grained search space by shrinking and expanding the model width, to precisely ensure the latency constraint. Combining the above components, we propose Three-Freedom Neural Architecture Search (TF-NAS) to search accurate latency-constrained architectures. 
To summarize, our main contributions lie in four-folds:
\begin{itemize}
\item Motivated by rethinking the operation-level, the depth-level and the width-level search freedoms, a novel TF-NAS is proposed to search accurate architectures with latency constraint.
\item We introduce a simple bi-sampling search algorithm to moderate operation collapse phenomenon. Besides, the mutual exclusion between skip and other candidate operations, as well as the architecture redundancy, are first considered to design a new sink-connecting search space. Both of them ensure the search flexibility and stability.
\item By investigating the coarse-grained discreteness of search space, we propose an elasticity-scaling strategy that progressively shrinks and expands the model width to ensure the latency constraint in a fine-grained manner.
\item Our TF-NAS can search architectures with precise latency on target devices, achieving state-of-the-art performance on ImageNet classification task. Particularly, our searched TF-NAS-A achieves 76.9\% top-1 accuracy with only 1.8 GPU days of search time.
\end{itemize}

\section{Related Work}

\textbf{Micro Search} focuses on finding robust cells~\cite{DBLP:conf/icml/PhamGZLD18,DBLP:conf/aaai/RealAHL19,DBLP:conf/icml/RealMSSSTLK17,DBLP:conf/iclr/XieZLL19,DBLP:conf/cvpr/ZophVSL18} and stacking many copies of them to design the network architecture. AmoebaNet~\cite{DBLP:conf/aaai/RealAHL19} and NASNet~\cite{DBLP:conf/cvpr/ZophVSL18}, which are based on Evolutionary Algorithm (EA) and  Reinforcement Learning (RL) respectively, are the pioneers of micro search algorithms. However, these approaches take an expensive computational overhead, i.e. over 2,000 GPU days, for searching. 
DARTS~\cite{DBLP:conf/iclr/LiuSY19} achieves a remarkable efficiency improvement (about 1 GPU day) by formulating the neural architecture search tasks in a differentiable manner. Following gradient based optimization in DARTS, 
GDAS~\cite{DBLP:conf/cvpr/DongY19} is proposed to sample one sub-graph from the whole directed acyclic graph (DAG) in one iteration, accelerating the search procedure. Xu et al.~\cite{DBLP:conf/iclr/XuXZCQTX20} randomly sample a proportion of channels for operation search in cells, leading to both faster search speed and higher training stability. P-DARTS~\cite{DBLP:conf/iccv/ChenXWT19} allows the depth of architecture to grow progressively in the search procedure, to alleviate memory/computational overheads and weak search instability. 
Comparing with accuracy, it is obvious that micro search algorithms are unfriendly to constrain the number of parameters, FLOPs and latency for neural architecture search.

\textbf{Macro Search} aims to search the entire neural architecture~\cite{DBLP:conf/iclr/CaiZH19,DBLP:journals/corr/abs-1908-06022,DBLP:conf/cvpr/TanCPVSHL19,DBLP:conf/icml/TanL19,DBLP:conf/cvpr/WuDZWSWTVJK19,DBLP:conf/iclr/ZophL17}, which is more flexible to obtain efficient networks. Baker et al.~\cite{DBLP:conf/iclr/BakerGNR17} introduce MetaQNN to sequentially choose CNN layers using Q-learning with an $\epsilon$-greedy exploration strategy. MNASNet~\cite{DBLP:conf/cvpr/TanCPVSHL19} and FBNet~\cite{DBLP:conf/cvpr/WuDZWSWTVJK19} are proposed to search efficient architectures with higher accuracy but lower latency. 
One-shot architecture search~\cite{DBLP:conf/icml/BenderKZVL18} designs a good search space and incorporates path drop when training the over-parameterized network. Since it suffers from the large memory usage to train an over-parameterized network, Cai et al.~\cite{DBLP:conf/iclr/CaiZH19} propose ProxylessNAS to provide a new path-level pruning perspective for NAS. 
Different from the previous neural architecture search, EfficientNet~\cite{DBLP:conf/icml/TanL19} proposes three model scaling factors including width, depth and resolution for network designment. Benefiting from compounding scales, they achieve state-of-the-art performance on various computer vision tasks. Inspired by EfficientNet~\cite{DBLP:conf/icml/TanL19}, in order to search for flexible architectures, we rethink three search freedoms, including \textbf{operation-level}, \textbf{depth-level} and \textbf{width-level}, for latency-constrained differentiable neural architecture search.

\section{Our Method}
\subsection{Review of Differentiable NAS}
\label{review_DNAS}
In this paper, we focus on differentiable neural architecture search to search accurate macro architectures constrained by various inference latencies. Similar with~\cite{DBLP:conf/cvpr/DongY19,DBLP:conf/iclr/LiuSY19,DBLP:conf/cvpr/WuDZWSWTVJK19}, the search problem is formulated as a bi-level optimization:
\begin{equation}
\mathop {\min }\limits_{\alpha  \in A} {L_{\text{val}}}\left( {{\omega ^*},\alpha } \right) + \lambda C\left( {LAT(\alpha )} \right)
\label{eq1}
\end{equation}
\begin{equation}
\text{s.t.}\quad{\omega ^*} = \mathop {\arg \min }\limits_\omega  {L_{\text{train}}}\left( {\omega ,\alpha } \right)
\label{eq2}
\end{equation}
where $\omega$ and $\alpha$ are the supernet weights and the architecture distribution parameters, respectively. Given a supernet $A$, we aim to search a subnet ${\alpha ^*} \in A$ that minimizes the validation loss ${L_{val}}\left( {{\omega ^*},\alpha } \right)$ and the latency constraint $C\left( {LAT(\alpha )} \right)$, where the weights ${{\omega ^*}}$ of supernet are obtained by minimizing the training loss ${L_{train}}\left( {\omega ,\alpha } \right)$ and $\lambda$ is a trade-off hyperparameter.

Different from RL-based~\cite{DBLP:conf/cvpr/TanCPVSHL19,DBLP:conf/iclr/ZophL17,DBLP:conf/cvpr/ZophVSL18} or EA-based~\cite{DBLP:journals/corr/abs-1908-06022,DBLP:conf/cvpr/DaiZWYSWDHWJVUJ19,DBLP:conf/aaai/RealAHL19} NAS, where the outer objective Eq.~(\ref{eq1}) is treated as reward or fitness, differentiable NAS optimizes Eq.~(\ref{eq1}) by gradient descent. Sampling a subnet from supernet $A$ is a non-differentiable process w.r.t. the architecture distribution parameters $\alpha$. Therefore, a continuous relaxation is needed to allow back-propagation. Assuming there are $N$ operations to be searched in each layer, we define ${\rm{op}}_i^l$ and $\alpha _i^l$ as the $i$-th operation in layer $l$ and its architecture distribution parameter, respectively. Let ${x^l}$ present the input feature map of layer $l$. A commonly used continuous relaxation is based on Gumbel Softmax trick~\cite{DBLP:conf/cvpr/DongY19,DBLP:conf/cvpr/WuDZWSWTVJK19}:
\begin{equation}
{x^{l + 1}} = \sum\limits_i {u_i^l \cdot } {\rm{op}}_i^l\left( {{x^l}} \right), u_i^l = \frac{{\exp \left( {(\alpha _i^l + g_i^l)/\tau } \right)}}{{\sum\limits_j {\exp \left( {(\alpha _j^l + g_j^l)/\tau } \right)} }}
\label{eq3}
\end{equation}
\begin{equation}
LAT(\alpha ) = \sum\limits_l {LAT\left( {{\alpha ^l}} \right)}  = \sum\limits_l {\sum\limits_i {u_i^l \cdot } } LAT\left( {{\rm{op}}_i^l} \right)
\label{eq4}
\end{equation}
where $\tau$ is the temperature parameter, $g_i^l$ is a random variable i.i.d sampled from $Gumbel(0,1)$, ${LAT\left( {{\alpha ^l}} \right)}$ is the latency of layer $l$ and $LAT\left( {{\rm{op}}_i^l} \right)$ is indexed from a pre-built latency lookup table. The superiority of Gumbel Softmax relaxation is to save GPU memory by approximate $N$ times and to reduce search time. That is because only one operation with max $u_i^l$ is chosen during forward pass. And the gradients of all the $\alpha _i^l$ can be back-propagated through Eq.~(\ref{eq3}).

\subsection{The Search Space}
\begin{figure*}[t]
  \centering
  \includegraphics[width=1.0\linewidth]{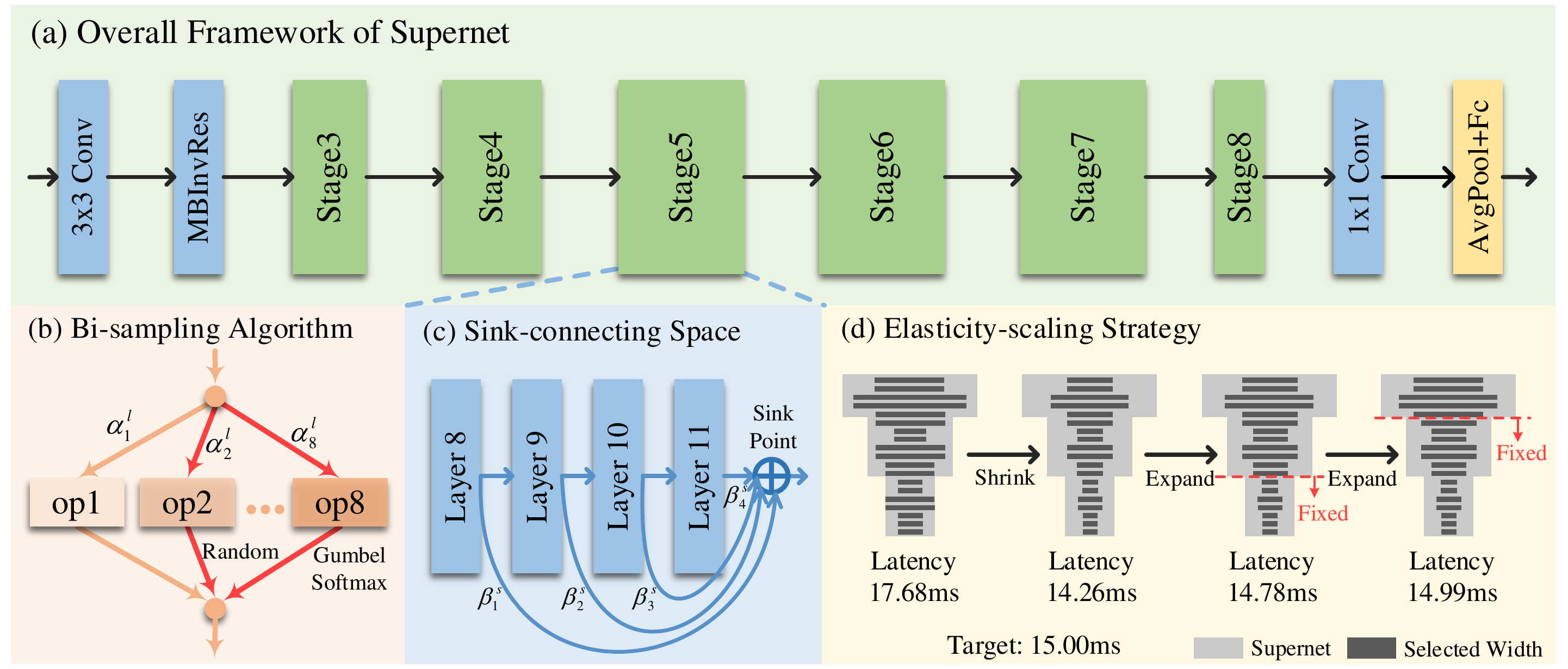} \vspace{-0.6cm}
  \caption{The search space of TF-NAS. It contains (b) operation-level, (c) depth-level and (d) width-level search freedoms.}
  \label{fig:overall_framwork}
\end{figure*}

In this paper, we focus on latency-constrained macro search. Inspired by EfficientNet~\cite{DBLP:conf/icml/TanL19}, we build a layer-wise search space, which is depicted in Fig.~\ref{fig:overall_framwork} and Tab.~\ref{tab:macro_arch_and_ops}. 
The input shapes and the channel numbers are the same as EfficientNet-B0~\cite{DBLP:conf/icml/TanL19}. Different from EfficientNet-B0, we use ReLU in the first three stages. The reason is that the large resolutions of the early inputs mainly dominate the inference latency, leading to worse optimization during architecture searching.

\begin{table*}[t]
\begin{minipage}{0.48\linewidth}
\centering
\resizebox{0.8\textwidth}{!}{
\begin{tabular}{c|c|c|c|c|c} \hline
  Stage & Input & Operation & $C_{out}$ & Act & L \\
  \hline
  1  & $224^2\times3$  & $3\times3$ Conv & 32 & ReLU & 1 \\
  2  & $112^3\times32$ & MBInvRes & 16 & ReLU & 1 \\
  3  & $112^2\times16$ & OPS & 24  & ReLU  & $[1, 2]$ \\
  4  & $56^2\times24$  & OPS & 40  & Swish & $[1, 3]$ \\
  5  & $28^2\times40$  & OPS & 80  & Swish & $[1, 4]$ \\
  6  & $14^2\times80$  & OPS & 112 & Swish & $[1, 4]$ \\
  7  & $14^2\times112$ & OPS & 192 & Swish & $[1, 4]$ \\
  8  & $7^2\times192$  & OPS & 320 & Swish & 1 \\
  9  & $7^2\times320$  & $1\times1$ Conv & 1280 & Swish & 1 \\
  10 & $7^2\times1280$ & AvgPool & 1280 & - & 1 \\
  11 & $1280$ & Fc & 1000 & - & 1 \\
  \hline
  \end{tabular}}
\end{minipage}
\begin{minipage}{0.48\linewidth}
\centering
\resizebox{0.8\textwidth}{!}{
\begin{tabular}{l|c|c|c} \hline
  OPS & Kernel & Expansion & SE Expansion \\
  \hline
  $k3\_e3$          & 3 & $[2, 4]$ & - \\
  $k3\_e3\_e_{se}1$ & 3 & $[2, 4]$ & 1 \\
  $k5\_e3$          & 5 & $[2, 4]$ & - \\
  $k5\_e3\_e_{se}1$ & 5 & $[2, 4]$ & 1 \\
  $k3\_e6$          & 3 & $[4, 8]$ & - \\
  $k3\_e6\_e_{se}2$ & 3 & $[4, 8]$ & 2 \\
  $k5\_e6$          & 5 & $[4, 8]$ & - \\
  $k5\_e6\_e_{se}2$ & 5 & $[4, 8]$ & 2 \\
  \hline
  \end{tabular}}
\end{minipage}
\caption{\textbf{Left}: Macro architecture of the supernet. ``OPS" denotes the operations to be searched. ``MBInvRes" is the basic block in~\cite{DBLP:conf/cvpr/SandlerHZZC18}. ``$C_{out}$" means the output channels. ``Act" denotes the activation function used in a stage. ``L" is the number of layers in a stage, where $[a, b]$ is a discrete interval. If necessary, the down-sampling occurs at the first operation of a stage. \textbf{Right}: Candidate operations to be searched. ``Expansion" defines the width of an operation and $[a, b]$ is a continuous interval. ``SE Expansion" determines the width of the SE module.} \vspace{-0.6cm}
\label{tab:macro_arch_and_ops}
\end{table*}

Layers from stage 3 to stage 8 are searchable, and each layer can choose an operation to form the operation-level search space. The basic units of the candidate operations are MBInvRes (the basic block in MobileNetV2~\cite{DBLP:conf/cvpr/SandlerHZZC18}) with or without Squeeze-and-Excitation (SE) module, which are illustrated in Appendix~\ref{details_of_MBInvRes_w_wo_SE_module}. 
In our experiments, there are 8 candidate operations to be searched in each searchable layer. The detailed configurations are listed in Tab.~\ref{tab:macro_arch_and_ops}. Each candidate operation has a kernel size $k=3$ or $k=5$ for the depthwise convolution, and a continuous expansion ratio $e \in [2, 4]$ or $e \in [4, 8]$, which constitutes to the width-level search space. 
Considering the operations with SE module, the SE expansion ratio is $e_{se}=1$ or $e_{se}=2$. 
In Tab.~\ref{tab:macro_arch_and_ops}, the ratio of $e_{se}$ to $e$ for all the candidate operations lies in $[0.25, 0.5]$. $e3$ or $e6$ in the first column of Tab.~\ref{tab:macro_arch_and_ops} defines the expansion ratio is $3$ or $6$ at the beginning of searching, and $e$ can vary in $[2, 4]$ or $[4, 8]$ during searching. Following the same naming schema, MBInvRes at stage 2 has a fixed configuration of $k3\_e1\_e_{se}0.25$. 
Besides, we also construct a depth-level search space based on a new sink-connecting schema. As shown in Fig.~\ref{fig:overall_framwork}(c), during searching, the outputs of all the layers in a stage are connected to a sink point, which is the input to the next stage. After searching, only one connection, i.e. depth, is chosen in each stage.

\subsection{Three-Freedom NAS}
\vspace{-0.2cm}
In this section, we investigate the operation-level, depth-level and width-level search freedoms, respectively, and accordingly make considerable improvements of the search flexibility and stability. Finally, our Three-Freedom NAS is summarized at the end of section.

\begin{figure}[h]
  \centering
  \includegraphics[width=1.0\linewidth]{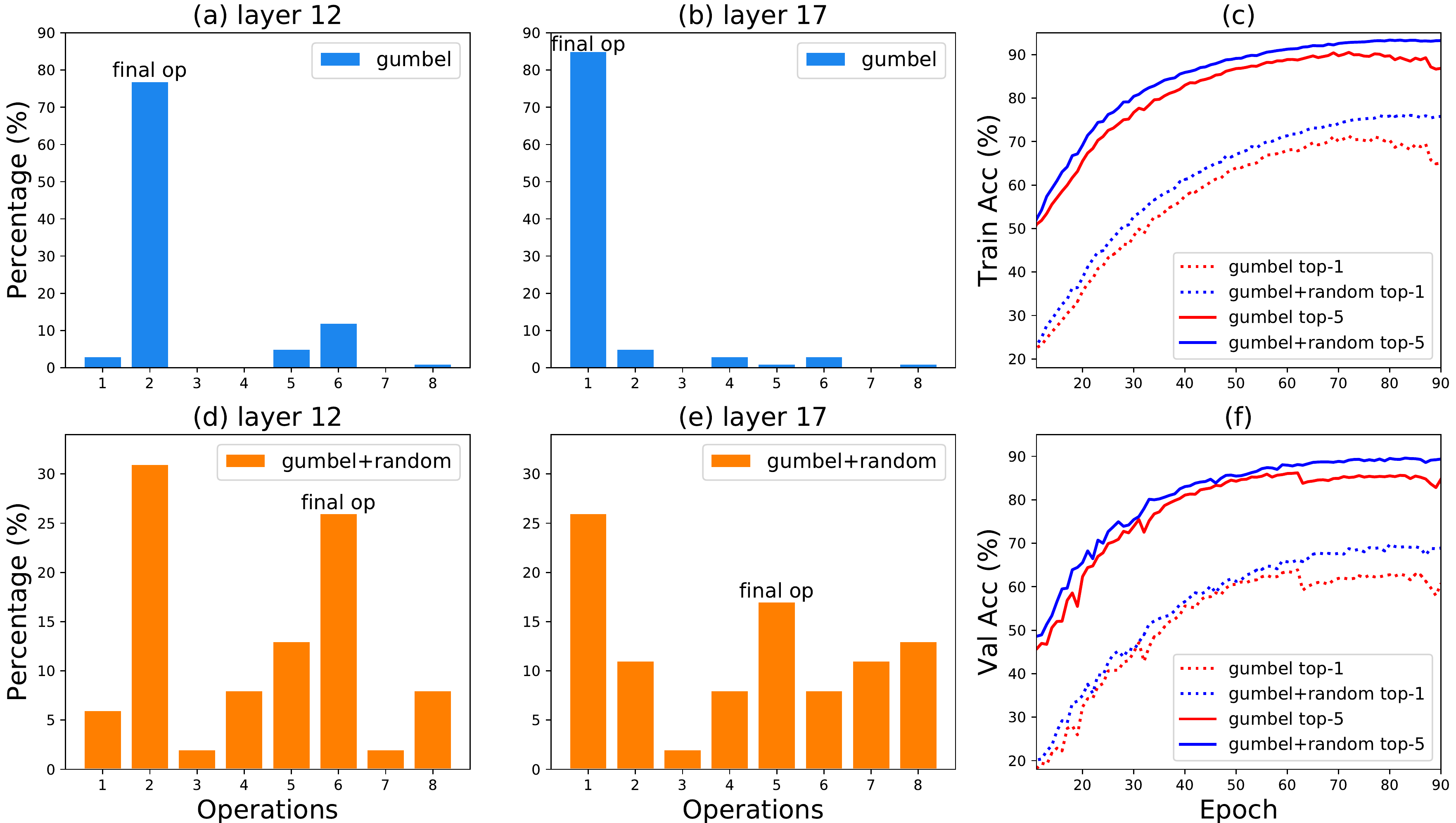} \vspace{-0.6cm}
  \caption{(a)-(b): The counting percentages of the derived operations during searching by Gumbel Softmax relaxation. (d)-(e): The counting percentages of the derived operations during searching by bi-sampling algorithm. (c): Training accuracy of the supernet. (f): Validating accuracy of the supernet. Zoom in for better view.} \vspace{-0.6cm}
  \label{fig:operation-level}
\end{figure}
\textbf{Rethinking Operation-level Freedom.}
\label{rethinking_operation-level_freedom}
As demonstrated in Sec.~\ref{review_DNAS}, NAS based on Gumbel Softmax relaxation samples one operation per layer during forward pass. It means when optimizing the inner objective Eq.~(\ref{eq2}), only one path is chosen and updated by gradient descent. However, due to the alternative update between $\omega$ and $\alpha$ in the bi-level optimization, one path sampling strategy may focus on some specific operations and update their parameters more frequently than others. Then the architecture distribution parameters of these operations will get better when optimizing Eq.~(\ref{eq1}). Accordingly, the same operation is more likely to be selected in the next sampling. This phenomenon may cause the search procedure to fall into the specific operations at some layers, leading to suboptimal architectures. We call it operation collapse. Although there is a temperature parameter $\tau$ to control the sampling, we find that the operation collapse still occurs in practice. We conduct an experiment based on our search space with the Gumbel Softmax relaxation, where $\tau$ linearly decreases from 5.0 to 0.2. The results are shown in Fig.~\ref{fig:operation-level}(a)-(b), where we count the derived operations for layer 12 and 17 during searching (after each search epoch). It can be observed that almost $80\%$ architecture derivations fall into specific operations in both layer 12 and 17, illustrating the occurrence of operation collapse.

To remedy the operation collapse, a straightforward method is early stopping~\cite{DBLP:journals/corr/abs-1909-06035,DBLP:conf/iccv/XiongMS19}. However, it may lead to suboptimal architectures due to incomplete supernet training and operation exploration (Appendix~\ref{comparison_with_early_stopping}). In this paper, we propose a simple bi-sampling search algorithm, where two independent paths are sampled for each time. In this way, when optimizing Eq.~(\ref{eq2}), two different paths are chosen and updated in a mini-batch. We implement it by conducting two times forward but one time backward. The second path is used to enhance the competitiveness of other operations against the one operation sampling in Gumbel Softmax. In Sec. \ref{bi-sampling_exp}, we conduct several experiments to explore various sampling strategies for the second path and find random sampling is the best one. Similarly, we also conduct an experiment based on our bi-sampling search algorithm and present the results in Fig.~\ref{fig:operation-level}(d)-(e). Compared with Gumbel Softmax based sampling, our bi-sampling strategy is able to explore more operations during searching. Furthermore, as shown in Fig.~\ref{fig:operation-level}(c) and Fig.~\ref{fig:operation-level}(f), our bi-sampling strategy is superior to the Gumbel Softmax based sampling in both the supernet accuracy on the training and the validating set.

\begin{figure}[t]
  \centering
  \includegraphics[width=1.0\linewidth]{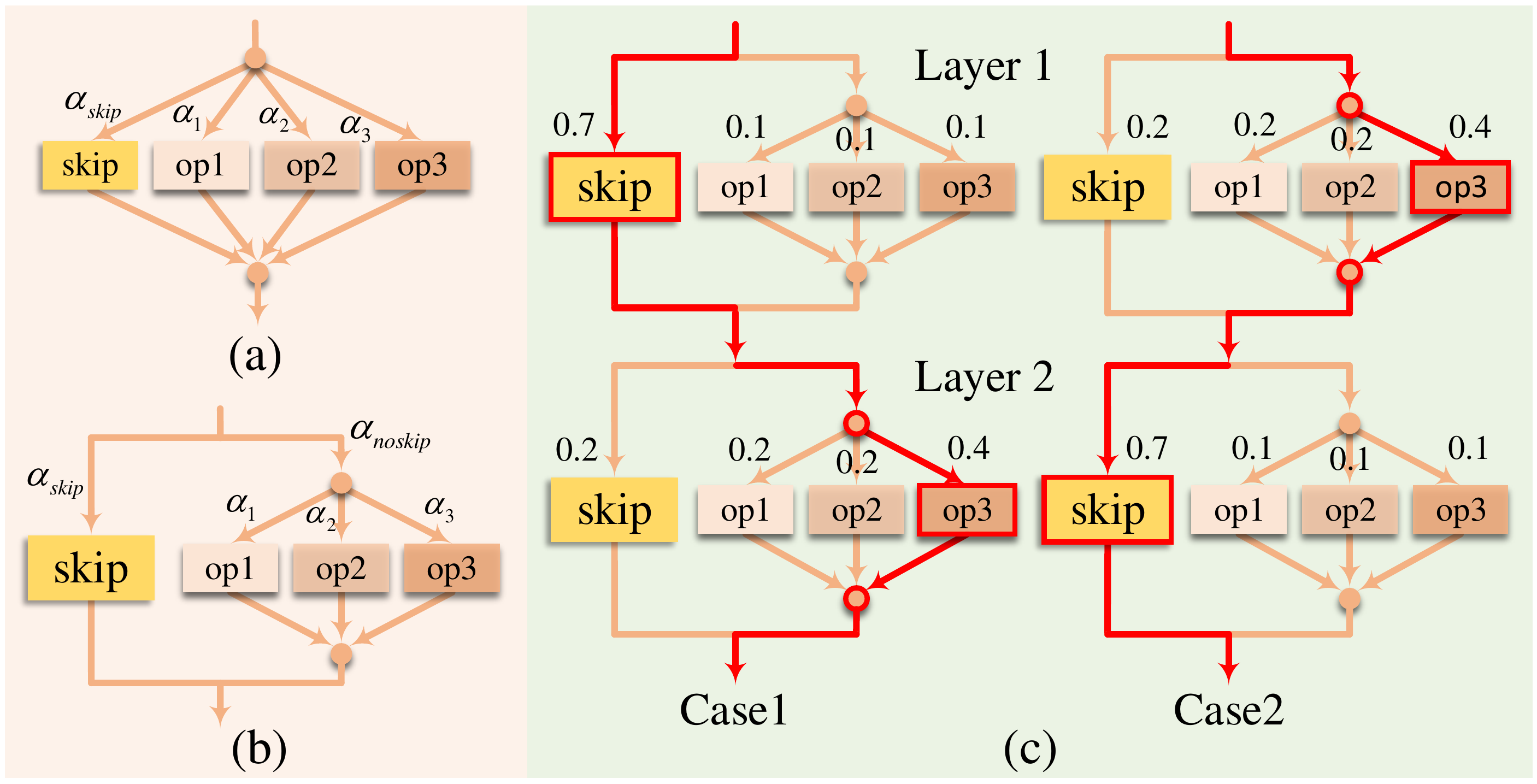} \vspace{-0.6cm}
  \caption{(a)-(b): The mutual exclusion between skip and other operations. (c): A case study for architecture redundancy.} \vspace{-0.6cm}
  \label{fig:depth_illustration}
\end{figure}
\begin{figure}[t]
  \centering
  \includegraphics[width=1.0\linewidth]{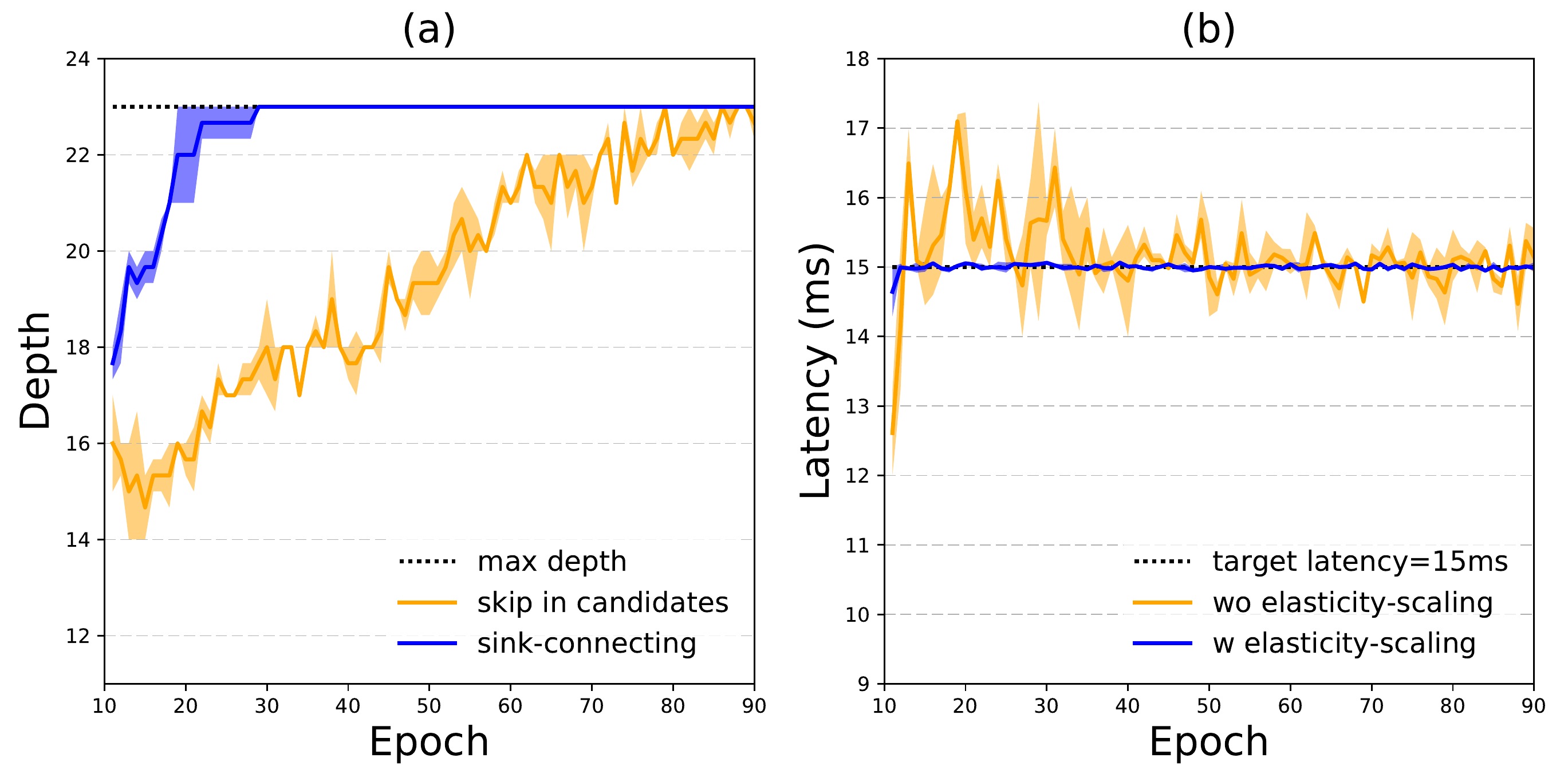} \vspace{-0.7cm}
  \caption{(a): The searched depth for different depth-level search spaces. (b): The searched latency w/wo elasticity-scaling.  All the search procedures are repeated 5 times, and we plot the mean, the maximum and the minimum. Zoom in for better view.} \vspace{-0.4cm}
  \label{fig:depth_and_width-level}
\end{figure}

\textbf{Rethinking Depth-level Freedom.}
\label{rethinking_depth-level_freedom}
In order to search for flexible architectures, an important component of differentiable NAS is depth-level search. Previous works~\cite{DBLP:journals/corr/abs-1908-06022,DBLP:conf/cvpr/WuDZWSWTVJK19} usually add a skip operation in the candidates and search them together (Fig.~\ref{fig:depth_illustration}(a)). In this case, skip has equal importance to other operations and the probability of $op2$ is $P(\alpha _2)$. However, it makes the search unstable, where the derived architecture is relatively shallow and the depth has a large jitter, especially in the early search phase, as shown in orange line in Fig.~\ref{fig:depth_and_width-level}(a). We argue that it is because the skip has higher priority to rule out other operations during searching, since it has no parameter. Therefore, the skip operation should be independent of other candidates, as depicted in Fig.~\ref{fig:depth_illustration}(b). We call it as the mutual exclusion between skip and other candidate operations. In this case, skip competes with all the other operations and the probability of $op2$ is $P(\alpha _2 ,\alpha _{noskip} )$. However, directly applying such a scheme will lead to architecture redundancy. Assuming there are two searchable layers in Fig.~\ref{fig:depth_illustration}(c). Case 1: we choose skip in layer 1 and $op3$ in layer 2. Case 2: we choose $op3$ in layer 1 and skip in layer 2. Both cases have the same derived architectures $op3$ but quite different architecture distributions. As the number of searchable layers increases, such architecture redundancy will be more serious.

To address the above issue, we introduce a new sink-connecting search space to ensure the mutual exclusion between skip and other candidate operations, as well as eliminate the architecture redundancy. The basic framework is illustrated in Fig.~\ref{fig:overall_framwork}(c), where the outputs of all the layers in a stage are connected to a sink point. During searching, the weighted sum of the output feature maps is calculated at the sink point, which is the input to the next stage. When deriving architectures, only one connection, i.e. depth, is chosen in each stage. Obviously, our sink-connecting search space makes the skip operation independent of the other candidates and has no architecture redundancy, because if a layer is skipped, then all the following layers in the same stage are also skipped. Let $\beta _l^s$ be the architecture distribution parameter of $l$-th connection in stage $s$. We employ a Softmax function as the continuous relaxation: \vspace{-0.2cm}

\begin{equation}
{x^{s + 1}} = \sum\limits_{l \in s} {v_l^s \cdot } {x^l},\;v_l^s = \frac{{\exp \left( {\beta _l^s} \right)}}{{\sum\limits_k {\exp \left( {\beta _k^s} \right)} }}
\label{eq5}
\end{equation}
\begin{equation}
Lat(\alpha ,\beta ) = \sum\limits_s {\sum\limits_{l \in s} {v_l^s  \cdot Lat(\alpha ^l )} }  = \sum\limits_s {\sum\limits_{l \in s} {\sum\limits_i {v_l^s  \cdot u_i^l  \cdot Lat(op_i^l )} } }
\label{eq6}
\end{equation} \vspace{-0.2cm}

Blue line in Fig.~\ref{fig:depth_and_width-level}(a) shows the search on sink-connecting search space. It is obvious that the search procedure is stable and the derived depth converges quickly. We do not sample for depth-level search, because if bi-sampling for $\beta$, we must independently sample 2 paths of depth and operation, respectively, leading to 4 times forward, which notably increases GPU memory and search time.

\textbf{Rethinking Width-level Freedom.}
Due to the coarse-grained discreteness of search space, current NAS methods cannot satisfy the precise latency constraints. Each searchable layer has a fixed number of channels for the candidate operations, which means each layer has a fixed number of latency options. Furthermore, in each stage, all the layers excluding the first one have the same input and output shapes, so the latency options of these layers are all the same. Although the search space of NAS is huge (e.g. it is $10^{21}$ for FBNet~\cite{DBLP:conf/cvpr/WuDZWSWTVJK19}), the statuses of architectures with different latencies are finite and discrete. Due to the coarse-grained search space for latency, some target latency cannot be precisely satisfied, leading to instability during architecture searching. For example, setting the target latency to be 15ms, we search two architectures: one is 14.32ms and the other is 15.76ms. Both of them have around 0.7ms gaps for the target latency. More analyses are presented in Sec.~\ref{elasticity-scaling_exp}.

In order to refine the coarse-grained search space for latency, previous works~\cite{DBLP:conf/cvpr/FangSZLLW20,DBLP:conf/cvpr/WuDZWSWTVJK19} introduce a global scaling factor or add additional candidate operations for width-level search. However, these methods are not flexible. Inspired by MorphNet~\cite{DBLP:conf/cvpr/GordonENCWYC18}, we propose an elasticity-scaling approach that adaptively shrinks and expands the model width to precisely satisfy the latency constraint in a progressively fine-grained manner. Our approach does not increase additional GPU memory and is insensitive to hyperparameter settings.

Given a supernet, we derive a discrete seed network ($sn$) based on the current architecture distribution parameters, where the strongest operation in each layer and the strongest depth in each stage are chosen. We can multiply $sn$ by a scaling factor $\gamma$ to control the width. Let $\gamma \cdot s{n_{i:j}}$ be a network whose layer width from stage $i$ to stage $j$ is multiplied by $\gamma$. Our elasticity-scaling strategy is presented in Algorithm \ref{algorithm_elasticity-scaling}, including a global scaling ($i=3$) and a series of progressively fine-grained scaling ($i=4 \dots 8$). Note that the searchable stages are from stage 3 to stage 8 in our search space. More implementation details can be found in Appendix~\ref{implementation_details_of_elasticity-scaling}. In Fig.~\ref{fig:depth_and_width-level} (b), we observe that our elasticity-scaling strategy is effective in stabilizing the architecture search with the precise latency constraint. \vspace{-0.7cm} 

\begin{algorithm}
\small
	\caption{Elasticity-scaling Strategy}
	\label{algorithm_elasticity-scaling}
	\begin{algorithmic}[1]
        \STATE Derive a seed network $sn$ from the supernet $A$.
        \STATE \textbf{for} {$i=3,\dots,8$} \textbf{do}
        \STATE ~~~Find the largest $\gamma$ such that $LAT(\gamma  \cdot s{n_{i:8}}) \le la{t_{\text{target}}}$.
        \STATE ~~~Set $sn = \gamma  \cdot s{n_{i:8}}$.
        \STATE \textbf{end for}
        \STATE Put $sn$ back to the supernet $A$.
        \STATE \textbf{return} $A$;
	\end{algorithmic}
\end{algorithm}
\vspace{-0.5cm}

\textbf{Overall Algorithm.}
\label{overall algorithm}
Our Three-Freedom NAS (TF-NAS) contains all above components: the bi-sampling search algorithm, the sink-connecting search space and the elasticity-scaling strategy. It finds latency-constrained architectures from the supernet (Tab.~\ref{tab:macro_arch_and_ops}) by solving the following bi-level problem:  \vspace{-0.1cm}
\begin{equation}
\mathop {\min }\limits_{\alpha ,\beta } {L_{\text{val}}}\left( {{\omega ^*},\alpha ,\beta } \right) + \lambda C\left( {LAT(\alpha ,\beta )} \right) \vspace{-0.2cm}
\label{eq7}
\end{equation}
\begin{equation}
\text{s.t.}\;{\omega ^*} = \mathop {\arg \min }\limits_\omega  {L_{\text{t\_g}}}\left( {\omega ,\alpha ,\beta } \right) + {L_{\text{t\_r}}}\left( {\omega ,\alpha ,\beta } \right) \vspace{-0.2cm}
\label{eq8}
\end{equation}
where ${L_{\text{t\_g}}}$ and ${L_{\text{t\_r}}}$ denote the training losses for Gumbel Softmax based sampling and random sampling, respectively. The latency-constrained objectives in~\cite{DBLP:conf/cvpr/WuDZWSWTVJK19,DBLP:conf/iclr/CaiZH19} do not employ the target latency, leading to imprecise latency compared with the target one. Therefore, we introduce a new objective that explicitly contains the target latency $la{t_{\text{target}}}$: \vspace{-0.2cm}
\begin{equation}
C\left( {LAT(\alpha ,\beta )} \right) = \max \left( {\frac{{LAT(\alpha ,\beta )}}{{la{t_{\text{target}}}}} - 1,0} \right)
\label{eq9}
\end{equation} \vspace{-0.3cm}

The continuous relaxations of $\alpha$ and $\beta$ are based on Eq.~(\ref{eq3})-(\ref{eq4}) and Eq.~(\ref{eq5})-(\ref{eq6}), respectively. We employ elasticity-scaling after each searching epoch, making it barely increase the search time. After searching, the best architecture is derived from the supernet based on $\alpha$ and $\beta$, where the strongest operation in each layer and the strongest depth in each stage are chosen.

\vspace{-0.3cm}
\section{Experiments}
\vspace{-0.2cm}
\subsection{Dataset and Settings}
\vspace{-0.2cm}
All the experiments are conducted on ImageNet~\cite{DBLP:conf/cvpr/DengDSLL009} under the mobile setting. Similar with~\cite{DBLP:conf/nips/CaiGH19}, the latency is measured with a batch size of 32 on a Titan RTX GPU. We set the number of threads for OpenMP to 1 and use Pytorch1.1+cuDNN7.6.0 to measure the latency. Before searching, we pre-build a latency look up table as described in~\cite{DBLP:conf/nips/CaiGH19,DBLP:conf/cvpr/WuDZWSWTVJK19}. To reduce the search time, we choose 100 classes from the original 1000 classes to train our supernet. The supernet is trained for 90 epochs, where the first 10 epochs do not update the architecture distribution parameters. This procedure takes about 1.8 days on 1 Titan RTX GPU. After searching, the derived architecture is trained from scratch on the whole ImageNet training set. For fair comparison, we train it for 250 epochs with standard data augmentation~\cite{DBLP:conf/iclr/CaiZH19}, in which no auto-augmentation or mixup is used. More experimental details are provided in Appendix~\ref{more_details_of_experimental_settings}.

\vspace{-0.2cm}
\subsection{Comparisons with Current SOTA}
\label{sota_exp}
\vspace{-0.2cm}
We compare TF-NAS with various manually designed and automatically searched architectures. According to the latency, we divide them into four groups. For each group, we set a target latency and search an architecture. Totally, there are four latency settings, including 18ms, 15ms, 12ms and 10ms, and the final architectures are named as TF-NAS-A, TF-NAS-B, TF-NAS-C and TF-NAS-D, respectively. The comparisons are presented in Tab.~\ref{tab:sota}. There is a slight latency error for each model. As shown in~\cite{DBLP:conf/iclr/CaiZH19}, the error mainly comes from the slight difference between the pre-built lookup table and the actual inference latency.

As shown in Tab.~\ref{tab:sota}, our TF-NAS-A achieves 76.9\% top-1 accuracy, which is better than NASNet-A~\cite{DBLP:conf/cvpr/ZophVSL18} (+2.9\%), PC-DARTS~\cite{DBLP:conf/iclr/XuXZCQTX20} (+1.1\%), MixNet-S~\cite{DBLP:conf/bmvc/TanL19} (+1.1\%) and EfficientNet-B0~\cite{DBLP:conf/icml/TanL19} (+0.6\%). For the GPU latency, 
TF-NAS-A is 6.2ms, 2.15ms, 1.83ms and 1.23ms better than NASNet-A, MdeNAS, PC-DARTS, MixNet-S and EfficientNet-B0, respectively. 
In the second group, our TF-NAS-B obtains 76.3\% top-1 accuracy with 15.06ms. 
It exceeds the micro search methods (DARTS~\cite{DBLP:conf/iclr/LiuSY19}, DGAS~\cite{DBLP:conf/cvpr/DongY19}, SETN~\cite{DBLP:conf/iccv/DongY19}, CARS-I~\cite{DBLP:conf/cvpr/YangWCSXXTX20}) by an average of 2.1\%, and the macro search methods (SCARLET-C~\cite{DBLP:journals/corr/abs-1908-06022}, DenseNAS-Large~\cite{DBLP:conf/cvpr/FangSZLLW20}) by an average of 0.5\%. For the 12ms latency group, our TF-NAS-C is superior to ShuffleNetV1 2.0x~\cite{DBLP:conf/cvpr/ZhangZLS18}, AtomNAS-A~\cite{DBLP:conf/iclr/MeiLLJYYY20}, FBNet-C~\cite{DBLP:conf/cvpr/WuDZWSWTVJK19} and ProxylessNAS (GPU)~\cite{DBLP:conf/iclr/CaiZH19} both in accuracy and latency. Besides, it is comparable with MobileNetV3~\cite{DBLP:conf/iccv/HowardSCCCTWZPVLA19} and MnasNet-A1~\cite{DBLP:conf/cvpr/TanCPVSHL19}. Note that MnasNet-A1 is trained for more epochs than our TF-NAS-C (350 vs 250). Obviously, training longer makes an architecture generalize better~\cite{DBLP:conf/nips/HofferHS17}. In the last group, our TF-NAS-D achieve 74.2\% top-1 accuracy, outperforming MobileNetV1~\cite{DBLP:journals/corr/HowardZCKWWAA17} (+3.6\%), ShuffleNetV1 1.5x~\cite{DBLP:conf/cvpr/ZhangZLS18} (+2.6\%) and FPNASNet~\cite{DBLP:conf/iccv/CuiCLLSJ19} (+0.9\%) by large margins.

Further to investigate the impact of the SE module, we remove SE from our candidate operations 
and search new architectures based on the four latency settings. The result architectures are marked as TF-NAS-A-wose, TF-NAS-B-wose, TF-NAS-C-wose and TF-NAS-D-wose. As shown in Tab.~\ref{tab:sota}, they obtain 76.5\%, 76.0\%, 75.0\% and 74.0\% top-1 accuracy, respectively, which are competitive with or even superior to the previous state-of-the-arts. Due to the page limitation, more results are presented in Appendix. \vspace{-0.6cm}

\begin{table}
  \centering
  \begin{tabular}{l|c|c|c|c|c|c} \hline
  Architecture & \tabincell{c}{Top-1\\Acc(\%)} & \tabincell{c}{GPU\\Latency} & \tabincell{c}{FLOPs\\(M)} & \tabincell{c}{Training\\Epochs} & \tabincell{c}{Search Time\\(GPU days)} & Venue \\
  \hline
  NASNet-A~\cite{DBLP:conf/cvpr/ZophVSL18}              & 74.0 & 24.23ms & 564 & -   & 2,000 & CVPR'18 \\
  PC-DARTS~\cite{DBLP:conf/iclr/XuXZCQTX20}             & 75.8 & 20.18ms & 597 & 250 & 3.8   & ICLR'20 \\
  MixNet-S~\cite{DBLP:conf/bmvc/TanL19}                 & 75.8 & 19.86ms & 256 & -   & -     & BMVC'19 \\
  EfficientNet-B0~\cite{DBLP:conf/icml/TanL19}          & 76.3 & 19.26ms & 390 & 350 & -     & ICML'19 \\
  TF-NAS-A-wose (Ours)                                  & 76.5 & 18.07ms & 504 & 250 & 1.8   & - \\
  TF-NAS-A (Ours)                                       & \textbf{76.9} & \textbf{18.03ms} & 457 & 250 & 1.8   & - \\
  \hline
  DARTS~\cite{DBLP:conf/iclr/LiuSY19}                   & 73.3 & 17.53ms & 574 & 250 & 4     & ICLR'19 \\
  DGAS~\cite{DBLP:conf/cvpr/DongY19}                    & 74.0 & 17.23ms & 581 & 250 & 0.21  & CVPR'19 \\
  SETN~\cite{DBLP:conf/iccv/DongY19}                    & 74.3 & 17.42ms & 600 & 250 & 1.8   & ICCV'19 \\
  MobileNetV2 1.4x~\cite{DBLP:conf/cvpr/SandlerHZZC18}  & 74.7 & 16.18ms & 585 & -   & -     & CVPR'18 \\
  CARS-I~\cite{DBLP:conf/cvpr/YangWCSXXTX20}            & 75.2 & 17.80ms & 591 & 250 & 0.4   & CVPR'20 \\
  SCARLET-C~\cite{DBLP:journals/corr/abs-1908-06022}    & 75.6 & 15.09ms & 280 & -   & 12    & ArXiv'19 \\
  DenseNAS-Large~\cite{DBLP:conf/cvpr/FangSZLLW20}      & 76.1 & 15.71ms & 479 & 240 & 2.67  & CVPR'20 \\
  TF-NAS-B-wose (Ours)                                  & 76.0 & 15.09ms & 433 & 250 & 1.8   & - \\
  TF-NAS-B (Ours)                                       & \textbf{76.3} & \textbf{15.06ms} & 361 & 250 & 1.8   & - \\
  \hline
  ShuffleNetV1 2.0x~\cite{DBLP:conf/cvpr/ZhangZLS18}    & 74.1 & 14.82ms & 524 & 240 & -     & CVPR'18 \\
  AtomNAS-A~\cite{DBLP:conf/iclr/MeiLLJYYY20}           & 74.6 & 12.21ms & 258 & 350 & -     & ICLR'20 \\
  FBNet-C~\cite{DBLP:conf/cvpr/WuDZWSWTVJK19}           & 74.9 & 12.86ms & 375 & 360 & 9     & CVPR'19 \\
  ProxylessNAS (GPU)~\cite{DBLP:conf/iclr/CaiZH19}      & 75.1 & 12.02ms & 465 & 300 & 8.3   & ICLR'18 \\
  MobileNetV3~\cite{DBLP:conf/iccv/HowardSCCCTWZPVLA19} & 75.2 & 12.36ms & 219 & -   & -     & ICCV'19 \\
  MnasNet-A1~\cite{DBLP:conf/cvpr/TanCPVSHL19}          & 75.2 & 11.98ms & 312 & 350 & 288   & CVPR'18 \\
  TF-NAS-C-wose (Ours)                                  & 75.0 & 12.06ms & 315 & 250 & 1.8   & - \\
  TF-NAS-C (Ours)                                       & \textbf{75.2} & \textbf{11.95ms} & 284 & 250 & 1.8   & - \\
  \hline
  MobileNetV1~\cite{DBLP:journals/corr/HowardZCKWWAA17} & 70.6 & \textbf{9.73ms}  & 569 & -   & -   & ArXiv'17 \\
  ShuffleNetV1 1.5x~\cite{DBLP:conf/cvpr/ZhangZLS18}    & 71.6 & 10.84ms & 292 & 240 & -     & CVPR'18 \\
  FPNASNet~\cite{DBLP:conf/iccv/CuiCLLSJ19}             & 73.3 & 11.60ms & 300 & -   & 0.83  & ICCV'19 \\
  TF-NAS-D-wose (Ours)                                  & 74.0 & 10.10ms & 286 & 250 & 1.8   & - \\
  TF-NAS-D (Ours)                                       & \textbf{74.2} & 10.08ms & 219 & 250 & 1.8   & - \\
  \hline
  \end{tabular}
  \caption{Comparisons with state-of-the-art architectures on the ImageNet classification task. For the competitors, we directly cite the FLOPs, the training epochs, the search time and the top-1 accuracy from their original papers or official codes. For the GPU latency, we measure it with a batch size of 32 on a Titan RTX GPU.}
  \label{tab:sota}
\end{table}

\subsection{Analyses of Bi-sampling Search Algorithm}
\label{bi-sampling_exp}
As described in Sec.~\ref{rethinking_operation-level_freedom}, our bi-sampling algorithm samples two paths in the forward pass. One path is based on Gumbel Softmax trick and the other is selected from the remaining paths. In this subsection, we set the target latency to 15ms and employ four types of sampling methods for the second path, including the Gumbel Softmax (Gumbel), the minimum architecture distribution parameter ($\min {\alpha ^l}$), the maximum architecture distribution parameter ($\max {\alpha ^l}$) and the random sampling (Random). As shown in Tab.~\ref{tab:bi-sampling}, compared with other methods, random sampling achieves the best top-1 accuracy. As a consequence, we employ random sampling in our bi-sampling search algorithm. Another interesting observation is that \textit{Gumbel+Gumbel} and \textit{Gumbel+$\max {\alpha ^l}$} are inferior to one path \textit{Gumbel} sampling strategy. This is due to the fact that both \textit{Gumbel+Gumbel} and \textit{Gumbel+$\max {\alpha ^l}$} will exacerbate the operation collapse phenomenon, leading to inferior architectures. Compared with one path \textit{Gumbel} sampling, our bi-sampling algorithm increases the search time by 0.3 GPU day, but makes a significant improvement in top-1 accuracy (76.3\% vs 75.8\%). 
\vspace{-0.2cm}

\begin{table}
  \centering
  \begin{tabular}{l|c|c|c|c} \hline
  Sampling & Top-1 Acc(\%) & GPU Latency & FLOPs(M) & Search Time \\
  \hline
  Gumbel                     & 75.8 & 15.05ms & 374 & 1.5 days \\
  Gumbel+Gumbel              & 75.7 & 15.04ms & 371 & 1.8 days \\
  Gumbel+$\min {\alpha ^l}$  & 76.0 & 15.11ms & 368 & 1.8 days \\
  Gumbel+$\max {\alpha ^l}$  & 75.5 & 14.92ms & 354 & 1.8 days \\
  Gumbel+Random              & 76.3 & 15.06ms & 361 & 1.8 days \\
  \hline
  \end{tabular}
  \caption{Comparisons with different sampling methods for the second path in bi-sampling search algorithm.} \vspace{-1.0cm}
  \label{tab:bi-sampling}
\end{table}

\begin{table}
  \centering
  \begin{tabular}{l|c|c|c|c|c} \hline
  Method & \tabincell{c}{Mutual\\Exclusion} & \tabincell{c}{Architecture\\Redundancy} & Top-1 Acc(\%) & GPU Latency & FLOPs(M) \\
  \hline
  skip in candidates  & $\times$ & $\times$ & 75.6 & 15.10ms & 384 \\
  skip out candidates & $\surd$  & $\times$ & 76.1 & 15.07ms & 376 \\
  sink-connecting     & $\surd$  & $\surd$  & 76.3 & 15.06ms & 361 \\
  \hline
  \end{tabular}
  \caption{Comparisons with different depth-level search spaces.} \vspace{-1.0cm}
  \label{tab:sink-connecting}
\end{table}

\subsection{Analyses of Sink-connecting Search Space}
\label{sink-connecting_exp}
As mentioned in Sec.~\ref{rethinking_depth-level_freedom}, skip operation has a special role in depth-level search. Ensuring the mutual exclusion between skip and other candidate operations, as well as eliminating the architecture redundancy are important stability factors for the architecture search procedure. In this subsection, we set the target latency to 15ms and compare our sink-connecting search space with the other two depth-level search spaces. The results are presented in Tab.~\ref{tab:sink-connecting}, where ``skip in candidates'' means adding the skip operation in the candidates (Fig.~\ref{fig:depth_illustration}(a)), and ``skip out candidates'' denotes putting the skip operation independent of the candidates (Fig.~\ref{fig:depth_illustration}(b)). Obviously, our ``sink-connecting'' achieves the best top-1 accuracy, demonstrating its effectiveness in finding accurate architectures during searching. The ``skip out candidates'' beats the ``skip in candidates'' by about 0.5\% top-1 accuracy, and the ``sink-connecting'' is 0.2\% higher than the ``skip out candidates''. The former achieves more improvement than the later, indicating that the mutual exclusion between skip and other operations is more important than the architecture redundancy. \vspace{-0.4cm}

\subsection{Analyses of Elasticity-scaling Strategy}
\label{elasticity-scaling_exp}
\begin{figure}
  \centering
  \includegraphics[width=1.0\linewidth]{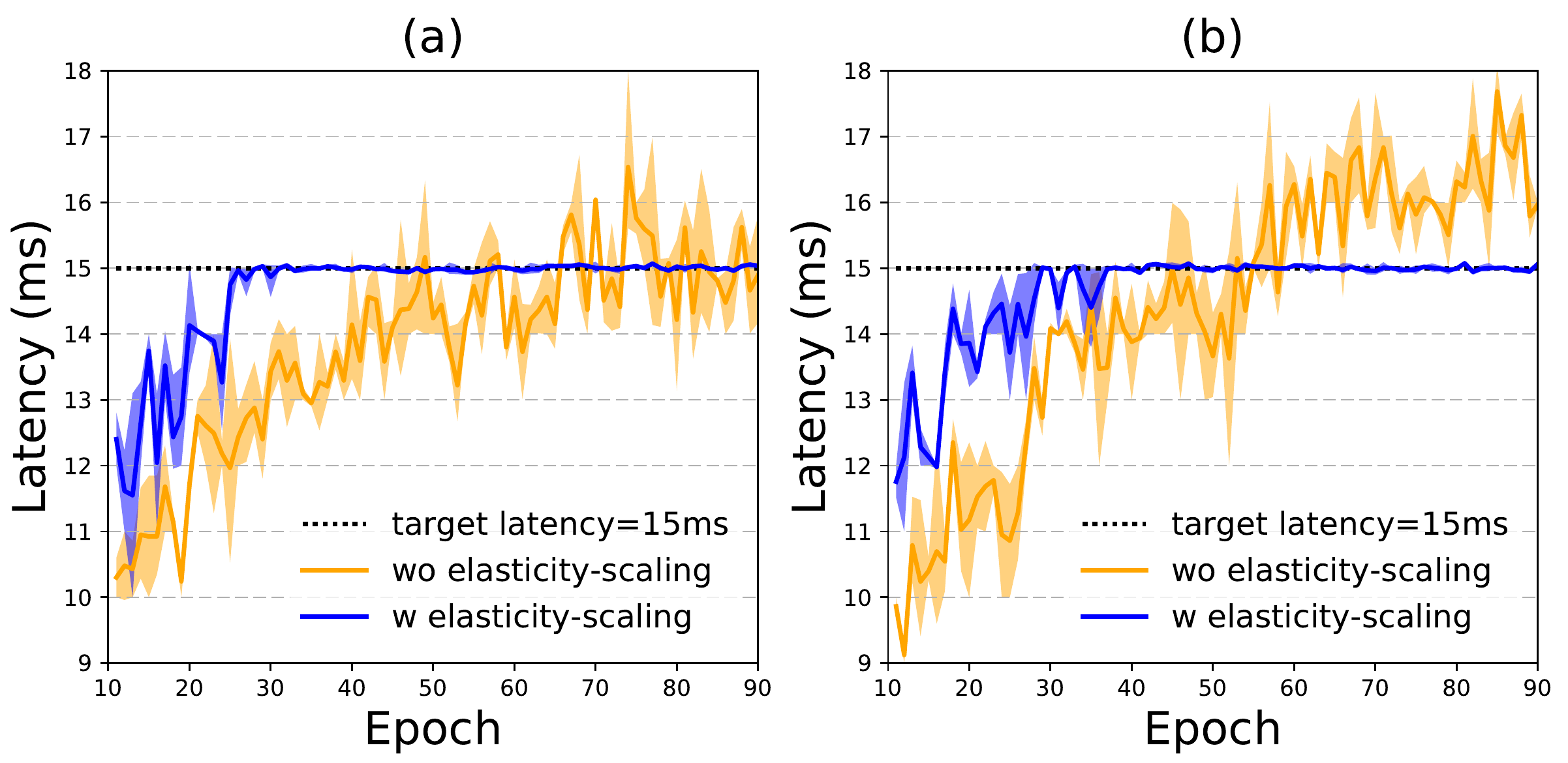}
  \caption{(a): The searched latency by C1 w/wo elasticity-scaling. (b): The searched latency by C2 w/wo elasticity-scaling. All the search procedures are repeated 5 times, and we plot the mean, the maximum and the minimum. Zoom in for better view.}
  \label{fig:lat_objs_compare}
\end{figure}

The key to search latency-constrained architectures is the differentiable latency objective $C\left( {LAT(\alpha ,\beta )} \right)$ in Eq.~(\ref{eq7}). Previous methods~\cite{DBLP:conf/cvpr/WuDZWSWTVJK19,DBLP:conf/iclr/CaiZH19} employ diverse latency objectives with one or two hyperparameters. We list them in Tab.~\ref{tab:elasticity-scaling} and name them as C1 and C2, respectively. By tuning the hyperparameters, both C1 and C2 can be trade-off between the accuracy and the latency. 
We set the target latency to 15ms and directly employ C1 and C2 (without elasticity-scaling strategy) to search architectures. We try our best to fine-tune the hyperparameters in C1 and C2, so that the searched architectures conform to the latency constraint as much as possible. The search procedure is repeated 5 times for each latency objective, and we plot the average latencies of the derived architectures during searching (orange lines in Fig.~\ref{fig:lat_objs_compare}(a)-(b)). It is obvious that both C1 and C2 cannot reach the target latency before the first 50 epochs. After that, the architecture searched by C1 fluctuates down and up around the target latency, but the architecture searched by C2 always exceeds the target latency. We also plot the results of our proposed latency objective Eq.~(\ref{eq9}) (orange line in Fig.~\ref{fig:depth_and_width-level}(b)) and find it is more precise than C1 and C2 after the first 30 epochs. The reason is that the target latency term is explicitly employed in our latency objective. 

The proposed elasticity-scaling strategy is the vital component in our TF-NAS to ensure the searched architectures precisely satisfy the target latency. By employing it, all the objectives are able to quickly search latency-satisfied architectures (blue lines in Fig.~\ref{fig:depth_and_width-level}(b), Fig.~\ref{fig:lat_objs_compare}(a) and Fig.~\ref{fig:lat_objs_compare}(b)), demonstrating the effectiveness and the versatility of our elasticity-scaling strategy. Furthermore, we also evaluate the searched architectures based on C1, C2 and our proposed objective with and without elasticity-scaling. As shown in Tab.~\ref{tab:elasticity-scaling}, our method achieves the best top-1 accuracy at 15ms latency constraint, which is slightly superior to C2 and beats C1 by a large margin no matter with or without elasticity-scaling. Therefore, explicitly introducing the target latency into the latency-constrained objective not only stabilizes large latency changes but also facilitates more accurate architecture discovery. Another observation is that under the similar backbone, the searched architectures with less/greater latencies than the target usually obtain lower/higher top-1 accuracies, especially when the latency gap is large. For example, C1 with elasticity-scaling achieves 75.9\% top-1/15.05ms, which beats its counterpart without elasticity-scaling (75.6\% top-1/14.32ms) by 0.3\% top-1 accuracy and the latency gap is approximate 0.7ms. 

\begin{table}
  \centering
  \begin{tabular}{l|c|c|c|c} \hline
  Name & Formulation & Elasticity-scaling & Top-1 Acc(\%) & GPU Latency \\
  \hline
  \multirow{2}{*}{C1~\cite{DBLP:conf/cvpr/WuDZWSWTVJK19}} & \multirow{2}{*}{${\lambda _1}\log {\left[ {\left( {LAT(\alpha ,\beta )} \right)} \right]^{{\lambda _2}}}$} & $\times$ & 75.6 & 14.32ms \\ \cline{3-5}
                                                          &                                                                                                            & $\surd$  & 75.9 & 15.05ms \\
  \hline
  \multirow{2}{*}{C2~\cite{DBLP:conf/iclr/CaiZH19}}       & \multirow{2}{*}{${\lambda _1}\left( {LAT(\alpha ,\beta )} \right)$} & $\times$ & 76.2 & 15.76ms \\ \cline{3-5}
                                                          &                                                                     & $\surd$  & 76.1 & 15.08ms \\
  \hline
  \multirow{2}{*}{Ours} & \multirow{2}{*}{${\lambda _1}\max \left( {\frac{{LAT(\alpha ,\beta )}}{{la{t_{\text{target}}}}} - 1,0} \right)$} & $\times$ & 76.3 & 15.28ms \\ \cline{3-5}
                        &                                                                                                                  & $\surd$  & 76.3 & 15.06ms \\
  \hline
  \end{tabular}
  \caption{Comparisons with different latency objectives w/wo elasticity-scaling.} \vspace{-1.2cm}
  \label{tab:elasticity-scaling}
\end{table}


\section{Conclusion}
In this paper, we have proposed Three-Freedom NAS (TF-NAS) to seek an architecture with good accuracy as well as precise latency on the target devices. For operation-level, the proposed bi-sample search algorithm moderates the operation collapse in Gumbel Softmax relaxation. For depth-level, a novel sink-connecting search space is defined to address the mutual exclusion between skip operation and other candidate operations, as well as architecture redundancy. For width-level, an elasticity-scaling strategy progressively shrinks or expands the width of operations, contributing to precise latency constraint in a fine-grained manner. Benefiting from investigating the three freedoms of differentiable NAS, our TF-NAS achieves state-of-the-art performance on ImageNet classification task. Particularly, the searched TF-NAS-A achieves 76.9\% top-1 accuracy with less latency and training epochs.

\subsubsection{Acknowledgement}
This work is partially funded by Beijing Natural Science Foundation (Grant No. JQ18017) and Youth Innovation Promotion Association CAS (Grant No. Y201929).

\clearpage
%
%
\bibliographystyle{splncs04}
\bibliography{egbib}

\appendix
\section{Details of MBInvRes w/wo SE Module}
\label{details_of_MBInvRes_w_wo_SE_module}
\begin{figure}
  \centering
  \includegraphics[width=1.0\linewidth,height=7cm]{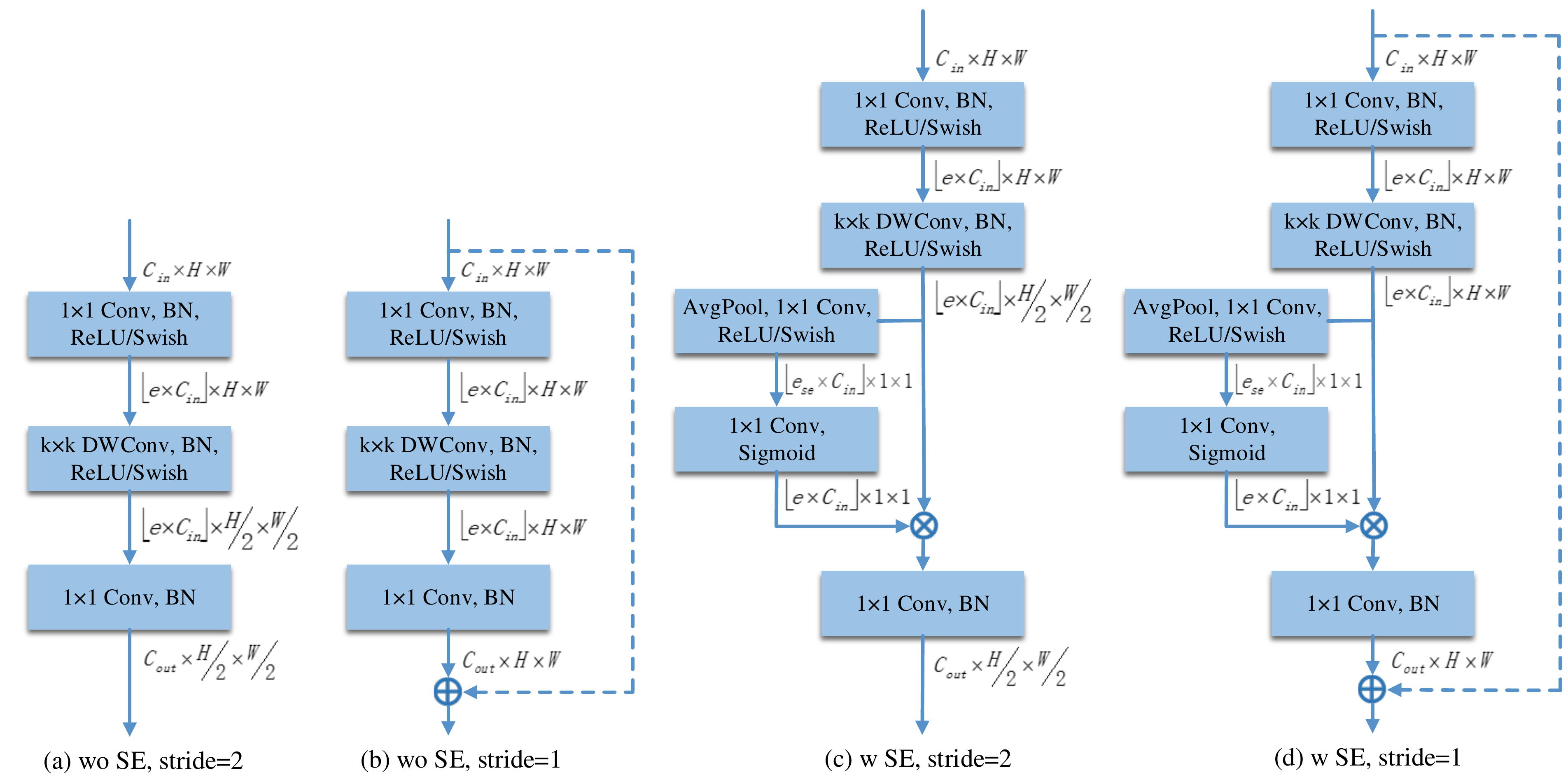}\vspace{-0.2cm}
  \caption{Illustrations of MBInvRes with or without SE module.}\vspace{-0.6cm}
  \label{fig:MBInvRes}
\end{figure}

The basic units of the candidate operations in our search space are MBInvRes with or without SE~\cite{DBLP:conf/cvpr/HuSS18} module. As illustrated in Fig.~\ref{fig:MBInvRes}, a MBInvRes without SE contains a point-wise convolution, followed by a $k \times k$ depthwise convolution and another point-wise convolution. Activation functions (ReLU or Swish) are equipped with the first point-wise convolution and the depthwise convolution, but not the last point-wise convolution. If the output shape is same as the input shape, we add a skip connection from the input to the output. As for the MBInvRes with SE, according to~\cite{DBLP:conf/iccv/HowardSCCCTWZPVLA19,DBLP:conf/icml/TanL19}, we put the SE module on the depthwise convolution, where a SE module consists of an average pooling, two fully connected layers and a sigmoid function.
\vspace{-0.2cm}

\section{More Details of Experimental Settings}
\label{more_details_of_experimental_settings}
In this section, we describe more details of experimental settings to facilitate other researchers to reproduce our results.

\textbf{Dataset.} All the experiments are conducted on the ImageNet~\cite{DBLP:conf/cvpr/DengDSLL009} dataset, which is a well-known and large-scale image classification benchmark. It totally contains 1.28 million images of 1,000 classes for training, and 50K images for validation. We employ the mobile setting in this paper, where the size of input images is $224 \times 224$ and the number of multiply-add operations is less than 600M.

\textbf{Latency Measurement.} Similar with~\cite{DBLP:conf/nips/CaiGH19}, the latency is measured with a batch size of 32 on a Titan RTX GPU. We set the number of threads for OpenMP to 1 and use Pytorch1.1+cuDNN7.6.0 to measure the latency. Before searching, we pre-build a latency look up table as described in~\cite{DBLP:conf/nips/CaiGH19,DBLP:conf/cvpr/WuDZWSWTVJK19}.

Our TF-NAS consists of two stages: architecture search and architecture evaluation. In architecture search, we train the supernet (Tab.~\ref{tab:macro_arch_and_ops} in the main text) on the ImageNet training set to find optimal architecture distribution parameters. In architecture evaluation, we derive the best architecture from the distribution parameters and train it from scratch.

\textbf{Architecture Search.} Similar with~\cite{DBLP:conf/cvpr/WuDZWSWTVJK19}, our supernet is trained for 90 epochs with a batch size of 32, where the first 10 epochs do not update the architecture distribution parameters $\alpha$ and $\beta$ to allow the supernet weights $\omega$ to be sufficiently trained first. To reduce the search time, we choose 100 classes from the original 1,000 classes to train our supernet. Instead of randomly sampling 100 classes as in~\cite{DBLP:conf/cvpr/FangSZLLW20,DBLP:conf/cvpr/WuDZWSWTVJK19}, we first employ a pre-trained EfficientNet-B0~\cite{DBLP:conf/icml/TanL19} to classify all the training images in ImageNet and calculate the top-1 accuracy of each class. Secondly, we resort the original 1,000 classes according to their accuracies and divide them into 100 groups. For each group, we randomly select one class to form the training set for our supernet. The supernet weights $\omega$ are trained on 80\% of the training set by SGD. We set the initial learning rate to 0.025 and anneal it down to zero by a cosine decaying schedule. The momentum is 0.9, and the weight decay is 1e-5. For the architecture distribution parameters $\alpha$ and $\beta$, we train them on the remaining 20\% of the training set by Adam. The learning rate, momentum and weight decay are set to 0.01, (0.5, 0.999) and 5e-4, respectively. We apply alternative optimization strategy to solve the bi-level optimization problem (Eq.~\ref{eq7}-\ref{eq8} in the main text). The temperature parameter $\tau$ is initially set to 5.0 and annealed by a factor of 0.96 for each epoch after the first 10 epochs. Besides, the trade-off parameter $\lambda$ is set to 0.1 in our experiments. We employ standard data augmentation~\cite{DBLP:conf/cvpr/HeZRS16} to train our supernet. The architecture search procedure takes about 1.8 days on 1 Titan RTX GPU.

\textbf{Architecture Evaluation.} After the supernet training, we derive the best architecture from the final architecture distribution parameters $\alpha^{*}$ and $\beta^{*}$, where the strongest operation in each layer and the strongest depth in each stage are chosen. The strengths of operations and depths are formulated as:
\begin{equation}
operation\_strength_i^l = \frac{{\exp \left( {\alpha _i^{*l}} \right)}}{{\sum\limits_j {\exp \left( {\alpha _j^{*l}} \right)} }}\vspace{-0.2cm}
\label{eq10}
\end{equation}
\begin{equation}
depth\_strength_l^s = \frac{{\exp \left( {\beta _l^{*s}} \right)}}{{\sum\limits_k {\exp \left( {\beta _k^{*s}} \right)} }}\vspace{-0.2cm}
\label{eq11}
\end{equation}

The derived architecture is trained from scratch on the whole ImageNet training set and tested on the ImageNet validation set. We train it by SGD with a batch size of 512, a momentum of 0.9 and a weight decay of 1e-5. The initial learning rate is set to 0.2 and annealed down to zero by a cosine decaying schedule. For fair comparison, we train the architecture for 250 epochs with standard data augmentation~\cite{DBLP:conf/iclr/CaiZH19}, where no auto-augmentation~\cite{DBLP:conf/cvpr/CubukZMVL19}, mixup~\cite{DBLP:conf/iclr/ZhangCDL18}, random erase~\cite{DBLP:conf/aaai/ZhongZKLY20} or any other augmentation is used. Linear warm-up is applied for the first 5 epochs due to the large batch size and learning rate. We employ a label smooth of 0.1 and set the dropout rate to 0.2, 0.2, 0.2 and 0.1 for TF-NAS-A/TF-NAS-A-wose, TF-NAS-B/TF-NAS-B-wose, TF-NAS-C/TF-NAS-C-wose and TF-NAS-D/TF-NAS-D-wose, respectively.

\section{Implementation Details of Elasticity-scaling}
\label{implementation_details_of_elasticity-scaling}
\begin{figure}
  \centering
  \includegraphics[width=1.0\linewidth]{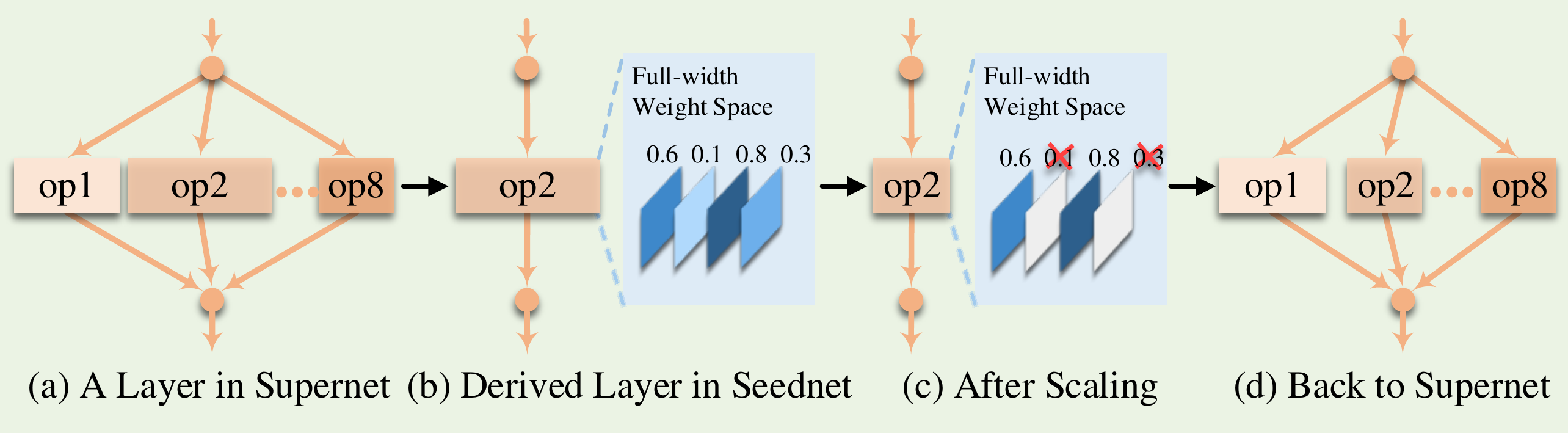}
  \caption{An example of shrinking an operation in the supernet.}\vspace{-0.6cm}
  \label{fig:width_illustration}
\end{figure}

Considering the detailed implementation of elasticity-scaling, we pre-allocate a full-width weight space for each candidate operation in the supernet. Once the width of an operation is changed by elasticity-scaling, we resort the channels according to their importance and choose the most important ones. The channel importance is calculated by the L1 norm of its corresponding weight. For example, when shrinking channels from $n$ to $m$ ($m<n$), we choose the top-$m$ channels whose weights are shared with the full-width weight space (Fig.~\ref{fig:width_illustration}(b)-(c)). Then, the shrunk operation is put back to the supernet, as shown in Fig.~\ref{fig:width_illustration}(d). If the same operation needs to be expanded in the future, the dropped channels can be reused. This weight sharing manner makes our approach no need to increase additional GPU memory.\vspace{-0.2cm}

\section{More Comparison Results}
Due to the page limitation, we only report some important methods in the main text Tab.~\ref{tab:sota}.
In this section, we compare our TF-NAS-A/B/C/D and TF-NAS-A/B/C/D-wose with more competitors under the mobile setting on ImageNet. The results are presented in Tab.~\ref{tab:more_sota}.

\begin{table}
  \centering
  \begin{tabular}{l|c|c|c|c|c|c} \hline
  Architecture & \tabincell{c}{Top-1\\Acc(\%)} & \tabincell{c}{GPU\\Latency} & \tabincell{c}{FLOPs\\(M)} & \tabincell{c}{Training\\Epochs} & \tabincell{c}{Search Time\\(GPU days)} & Venue \\
  \hline
  NASNet-A~\cite{DBLP:conf/cvpr/ZophVSL18}                    & 74.0 & 24.23ms & 564 & -   & 2,000 & CVPR'18 \\
  RCNet-B~\cite{DBLP:conf/iccv/XiongMS19}                     & 74.7 & 20.93ms & 471 & 400 & 8     & ICCV'18 \\
  MdeNAS~\cite{DBLP:conf/iccv/ZhengJTZLT19}                   & 75.2 & 18.65ms & 516 & 250 & 2     & ICCV'19 \\
  PC-DARTS~\cite{DBLP:conf/iclr/XuXZCQTX20}                   & 75.8 & 20.18ms & 597 & 250 & 3.8   & ICLR'20 \\
  MixNet-S~\cite{DBLP:conf/bmvc/TanL19}                       & 75.8 & 19.86ms & 256 & -   & -     & BMVC'19 \\
  SGAS (Cri.2)~\cite{DBLP:conf/cvpr/LiQDMTG20}                & 75.9 & 19.59ms & 598 & 250 & 0.25  & CVPR'20 \\
  XNAS~\cite{DBLP:conf/nips/NaymanNRFJZ19}                   & 76.0 & 18.86ms & 592 & 250 & 0.3   & NeurIPS'19 \\
  EfficientNet-B0~\cite{DBLP:conf/icml/TanL19}               & 76.3 & 19.26ms & 390 & 350 & -     & ICML'19 \\
  TF-NAS-A-wose (Ours)                                        & 76.5 & 18.07ms & 504 & 250 & 1.8   & - \\
  TF-NAS-A (Ours)                                             & \textbf{76.9} & \textbf{18.03ms} & 457 & 250 & 1.8   & - \\
  \hline
  DARTS~\cite{DBLP:conf/iclr/LiuSY19}                         & 73.3 & 17.53ms & 574 & 250 & 4     & ICLR'19 \\
  DGAS~\cite{DBLP:conf/cvpr/DongY19}                          & 74.0 & 17.23ms & 581 & 250 & 0.21  & CVPR'19 \\
  PNASNet-5~\cite{DBLP:conf/eccv/LiuZNSHLFYHM18}              & 74.2 & 16.04ms & 588 & 250 & 150   & ECCV'18 \\
  SETN~\cite{DBLP:conf/iccv/DongY19}                          & 74.3 & 17.42ms & 600 & 250 & 1.8   & ICCV'19 \\
  NAO~\cite{DBLP:conf/neurips/LuoTQL18}                       & 74.3 & 16.33ms & 584 & 250 & 24    & NeurIPS'18 \\
  BASE~\cite{DBLP:conf/nips/ShawWLSD19}                       & 74.3 & 16.19ms & 559 & -   & 8.04  & NeurIPS'19 \\
  MobileNetV2 1.4$\times$~\cite{DBLP:conf/cvpr/SandlerHZZC18} & 74.7 & 16.18ms & 585 & -   & -     & CVPR'18 \\
  CARS-I~\cite{DBLP:conf/cvpr/YangWCSXXTX20}                  & 75.2 & 17.80ms & 591 & 250 & 0.4   & CVPR'20 \\
  P-DARTS~\cite{DBLP:conf/iccv/ChenXWT19}                     & 75.6 & 17.79ms & 557 & 250 & 0.3   & ICCV'19 \\
  SCARLET-C~\cite{DBLP:journals/corr/abs-1908-06022}          & 75.6 & 15.09ms & 280 & -   & 12    & ArXiv'19 \\
  DenseNAS-Large~\cite{DBLP:conf/cvpr/FangSZLLW20}            & 76.1 & 15.71ms & 479 & 240 & 2.67  & CVPR'20 \\
  TF-NAS-B-wose (Ours)                                        & 76.0 & 15.09ms & 433 & 250 & 1.8   & - \\
  TF-NAS-B (Ours)                                             & \textbf{76.3} & \textbf{15.06ms} & 361 & 250 & 1.8   & - \\
  \hline
  SNAS (mild)~\cite{DBLP:conf/iclr/XieZLL19}                  & 72.7 & 12.61ms & 522 & 250 & 1.5   & ICLR'19 \\
  ShuffleNetV1 2.0$\times$~\cite{DBLP:conf/cvpr/ZhangZLS18}   & 74.1 & 14.82ms & 524 & 240 & -     & CVPR'18 \\
  AtomNAS-A~\cite{DBLP:conf/iclr/MeiLLJYYY20}                 & 74.6 & 12.21ms & 258 & 350 & -     & ICLR'20 \\
  FBNet-C~\cite{DBLP:conf/cvpr/WuDZWSWTVJK19}                 & 74.9 & 12.86ms & 375 & 360 & 9     & CVPR'19 \\
  SPOS~\cite{DBLP:conf/eccv/GuoZMHLWS20}                      & 74.9 & \textbf{11.89ms} & 328 & 240 & 12    & ECCV'20 \\
  ProxylessNAS (GPU)~\cite{DBLP:conf/iclr/CaiZH19}            & 75.1 & 12.02ms & 465 & 300 & 8.3   & ICLR'18 \\
  MobileNetV3~\cite{DBLP:conf/iccv/HowardSCCCTWZPVLA19}       & 75.2 & 12.36ms & 219 & -   & -     & ICCV'19 \\
  MnasNet-A1~\cite{DBLP:conf/cvpr/TanCPVSHL19}                & 75.2 & 11.98ms & 312 & 350 & 288   & CVPR'18 \\
  TF-NAS-C-wose (Ours)                                        & 75.0 & 12.06ms & 315 & 250 & 1.8   & - \\
  TF-NAS-C (Ours)                                             & \textbf{75.2} & 11.95ms & 284 & 250 & 1.8   & - \\
  \hline
  MobileNetV1~\cite{DBLP:journals/corr/HowardZCKWWAA17}       & 70.6 & \textbf{9.73ms}  & 569 & -   & -   & ArXiv'17 \\
  ShuffleNetV1 1.5$\times$~\cite{DBLP:conf/cvpr/ZhangZLS18}   & 71.6 & 10.84ms & 292 & 240 & -     & CVPR'18 \\
  MobileNetV2~\cite{DBLP:conf/cvpr/SandlerHZZC18}             & 72.0 & 11.15ms & 300 & -   & -     & CVPR'18 \\
  FPNASNet~\cite{DBLP:conf/iccv/CuiCLLSJ19}                   & 73.3 & 11.60ms & 300 & -   & 0.83  & ICCV'19 \\
  MobileNetV3 0.75x~\cite{DBLP:conf/iccv/HowardSCCCTWZPVLA19} & 73.3 & 10.01ms & 155 & -   & -     & ICCV'19 \\
  TF-NAS-D-wose (Ours)                                        & 74.0 & 10.10ms & 286 & 250 & 1.8   & - \\
  TF-NAS-D (Ours)                                             & \textbf{74.2} & 10.08ms & 219 & 250 & 1.8   & - \\
  \hline
  \end{tabular} \vspace{0.2cm}
  \caption{More comparison results under the mobile setting on the ImageNet classification task. For the competitors, we directly cite the FLOPs, the training epochs, the search time and the top-1 accuracy from their original papers or official codes.} 
  \label{tab:more_sota}
\end{table}

\section{Comparison with Early Stopping}
\vspace{-0.2cm}
\label{comparison_with_early_stopping}
In operation-level search, there is a straightforward method to remedy the operation collapse, i.e. early stopping. In this section, we compare our bi-sampling algorithm with the previous early stopping method~\cite{DBLP:journals/corr/abs-1909-06035}. For early stopping, we conduct a search by Gumbel sampling and Criterion 1* in~\cite{DBLP:journals/corr/abs-1909-06035}. Since we find there are several layers that cannot meet the original Criterion 1* during searching, we relax to stop when the ranking of architecture parameters for 3/4 layers becomes stable for 5 epochs. Setting the target to 15ms, we stop searching at the 64-th epoch and obtain 75.7\% top-1 accuracy. For fair comparison, we also evaluate the TF-NAS model derived from the 64-th search epoch and obtain 76.1\% top-1 accuracy, 0.4\% higher than early stopping. In fact, early stopping stops the search when collapse occurs, which is a way of “stop-losses” but cannot alleviate collapse.\vspace{-0.2cm}

\section{Transfer Learning on CIFAR10 and CIFAR100}
\vspace{-0.2cm}
Following EfficientNet~\cite{DBLP:conf/icml/TanL19}, we transfer the searched architectures TF-NAS-A, TF-NAS-B, TF-NAS-C and TF-NAS-D from ImageNet to CIFAR10~\cite{DBLP:journals/tr/KrizhevskyH09} and CIFAR100~\cite{DBLP:journals/tr/KrizhevskyH09} by resizing the images from $32\times32$ to $224\times224$. The results are shown in Tab.~\ref{tab:c10_and_c100}. \vspace{-0.4cm}

\begin{table}
  \small
  \centering
  \begin{tabular}{l|c|c|c} \hline
  Architecture & CIFAR10 Acc(\%) & CIFAR100 Acc(\%) & FLOPs(M) \\
  \hline
  TF-NAS-A  & 98.27 & 88.45 & 457 \\
  TF-NAS-B  & 98.13 & 88.26 & 361 \\
  TF-NAS-C  & 97.96 & 87.27 & 284 \\
  TF-NAS-D  & 97.78 & 85.83 & 219 \\
  \hline
  \end{tabular}\vspace{0.1cm}
  \caption{Transfer learning results on CIFAR10 and CIFAR100.}\vspace{-1.3cm}
  \label{tab:c10_and_c100}
\end{table}

\section{Searching for CPU Constrained Architectures}
\vspace{-0.2cm}
In this section, we demonstrate the results of architecture search with constraint of CPU latency. We measure the CPU latency via PyTorch1.1, with a batch size of 1 in single thread on Intel Xeon Gold 6130 @ 2.10GHz. Similarly, we pre-build a latency look up table as described in~\cite{DBLP:conf/nips/CaiGH19,DBLP:conf/cvpr/WuDZWSWTVJK19}. We make two latency settings of 60ms and 40ms, and named the searched architectures as TF-NAS-CPU-A and TF-NAS-CPU-B, respectively. All the search and evaluation hyperparameters are consistent with Appendix~\ref{more_details_of_experimental_settings}, except that the dropout rate of TF-NAS-CPU-A and TF-NAS-CPU-B are both set to 0.2.

As shown in Tab.~\ref{tab:cpu_constrained_comparison}, our TF-NAS-CPU-A achieves 75.8\% top-1 accuracy, outperforming MobileNetV2 1.4$\times$~\cite{DBLP:conf/cvpr/SandlerHZZC18} (+1.1\%), RCNet-B~\cite{DBLP:conf/iccv/XiongMS19} (1.1\%) and SPOS~\cite{DBLP:conf/eccv/GuoZMHLWS20} (0.9\%) by large margins with a similar CPU latency. Compared with ProxylessNAS (CPU)~\cite{DBLP:conf/iclr/CaiZH19}, TF-NAS-CPU-A reduces the CPU latency by about 30\% and improves the top-1 accuracy by 0.5. On pair with MixNet-S~\cite{DBLP:conf/bmvc/TanL19}, it further obtains 1.63$\times$ speed up on Intel Xeon Gold 6130 @ 2.10GHz. For the group of 40ms, our TF-NAS-CPU-B is superior to MobileNetV1~\cite{DBLP:journals/corr/HowardZCKWWAA17}, MobileNetV2~\cite{DBLP:conf/cvpr/SandlerHZZC18}, DenseNAS-A~\cite{DBLP:conf/cvpr/FangSZLLW20}, MobileNetV3 0.75$\times$~\cite{DBLP:conf/iccv/HowardSCCCTWZPVLA19} and FPNASNet~\cite{DBLP:conf/iccv/CuiCLLSJ19} on both the top-1 accuracy and the CPU latency. In addition, Tab.~\ref{tab:gpu_vs_cpu} presents all the searched TF-NAS models. Obviously, no matter on GPU or CPU, the actual inference latency is almost the same as the lookup table. It not only illustrates the effectiveness of the pre-built lookup table, but also demonstrates that our method is able to achieve precise latency constraint. \vspace{-0.4cm}

\begin{table}
  \centering
  \resizebox{0.85\textwidth}{!}{
  \begin{tabular}{l|c|c|c|c|c} \hline
  Architecture & \tabincell{c}{Top-1\\Acc(\%)} & \tabincell{c}{CPU\\Latency} & \tabincell{c}{FLOPs\\(M)} & \tabincell{c}{Training\\Epochs} & \tabincell{c}{Search Time\\(GPU days)} \\
  \hline
  MobileNetV2 1.4$\times$~\cite{DBLP:conf/cvpr/SandlerHZZC18}        & 74.7 & 75.11ms & 585 & -   & -    \\
  RCNet-B~\cite{DBLP:conf/iccv/XiongMS19}                            & 74.7 & 69.49ms & 471 & 400 & 8    \\
  SPOS~\cite{DBLP:conf/eccv/GuoZMHLWS20}                             & 74.9 & 60.92ms & 328 & 240 & 12   \\
  ProxylessNAS (CPU)~\cite{DBLP:conf/iclr/CaiZH19}                   & 75.3 & 84.81ms & 439 & 300 & 8.3  \\
  MixNet-S~\cite{DBLP:conf/bmvc/TanL19}                              & 75.8 & 97.92ms & 256 & -   & -    \\
  TF-NAS-CPU-A (Ours)                                                & \textbf{75.8} & \textbf{60.11ms} & 305 & 250 & 1.8  \\
  \hline
  MobileNetV1~\cite{DBLP:journals/corr/HowardZCKWWAA17}              & 70.6 & 44.93ms & 569 & -   & -    \\
  MobileNetV2~\cite{DBLP:conf/cvpr/SandlerHZZC18}                    & 72.0 & 55.46ms & 300 & -   & -    \\
  DenseNAS-A~\cite{DBLP:conf/cvpr/FangSZLLW20}                       & 73.1 & 40.21ms & 251 & 240 & 2.67 \\
  MobileNetV3 0.75$\times$~\cite{DBLP:conf/iccv/HowardSCCCTWZPVLA19} & 73.3 & 41.48ms & 155 & -   & -    \\
  FPNASNet~\cite{DBLP:conf/iccv/CuiCLLSJ19}                          & 73.3 & 42.41ms & 300 & -   & 0.83 \\
  TF-NAS-CPU-B (Ours)                                                & \textbf{74.4} & \textbf{40.09ms} & 230 & 250 & 1.8  \\
  \hline
  \end{tabular}}\vspace{0.2cm}
  \caption{Comparison results of CPU constrained TF-NAS with other manually or automatically designed architectures on the ImageNet classification task. The CPU latency is measured with a batch size of 1 on Intel Xeon Gold 6130 @ 2.10GHz.} \vspace{-1.4cm}
  \label{tab:cpu_constrained_comparison}
\end{table}

\begin{table}
  \centering
  \resizebox{0.85\textwidth}{!}{
  \begin{tabular}{l|c|c|c|c|c|c} \hline
  Architecture & \tabincell{c}{Top-1\\Acc(\%)} & \tabincell{c}{GPU\\Latency} & \tabincell{c}{GPU Lookup\\Table} & \tabincell{c}{CPU\\Latency} & \tabincell{c}{CPU Lookup\\Table} & \tabincell{c}{FLOPs\\(M)} \\
  \hline
  TF-NAS-A        & 76.9 & 18.03ms & 17.99ms & 80.14ms & -       & 457 \\
  TF-NAS-B        & 76.3 & 15.06ms & 14.99ms & 72.10ms & -       & 361 \\
  TF-NAS-C        & 75.2 & 11.95ms & 12.03ms & 51.87ms & -       & 284 \\
  TF-NAS-D        & 74.2 & 10.08ms &  9.99ms & 46.09ms & -       & 219 \\
  \hline
  TF-NAS-A-wose   & 76.5 & 18.07ms & 17.99ms & 72.67ms & -       & 504 \\
  TF-NAS-B-wose   & 76.0 & 15.09ms & 14.99ms & 67.66ms & -       & 433 \\
  TF-NAS-C-wose   & 75.0 & 12.06ms & 12.04ms & 49.29ms & -       & 315 \\
  TF-NAS-D-wose   & 74.0 & 10.10ms &  9.99ms & 44.86ms & -       & 286 \\
  \hline
  TF-NAS-CPU-A    & 75.8 & 14.00ms & -       & 60.11ms & 59.99ms & 305 \\
  TF-NAS-CPU-B    & 74.4 & 10.29ms & -       & 40.09ms & 40.18ms & 230 \\
  \hline
  \end{tabular}}\vspace{0.2cm}
  \caption{Comparisons between GPU and CPU constrained TF-NAS on ImageNet. The `GPU/CPU Lookup Table' means the latency is calculated from the pre-built lookup table.} \vspace{-1.4cm}
  \label{tab:gpu_vs_cpu}
\end{table}

\section{Results on MobileNetV2-based Search Space}
\vspace{-0.2cm}
In this section, we conduct several experiments on MobileNetV2~\cite{DBLP:conf/cvpr/SandlerHZZC18}-based search space to demonstrate the universality of TF-NAS. As shown in Tab.~\ref{tab:mbv2_macro_arch_and_ops}, the first two and the last three layers (stages) are fixed and the rest layers are searchable. There are total 4 candidate operations to be searched in each searchable layer, where the basic unit is MBInvRes~\cite{DBLP:conf/cvpr/SandlerHZZC18}. The detailed configurations are listed on the right side of Tab.~\ref{tab:mbv2_macro_arch_and_ops}. Each candidate operation has a kernel size $k=3$ or $k=5$ and a continuous expansion ratio $e \in [2, 4]$ or $e \in [4, 8]$. The MBInvRes at stage 2 has a fixed configuration of $k3\_e1$. We search for architectures based on GPU latency and make two latency settings: 15ms and 10ms. The searched architectures are named as TF-NAS-MBV2-A and TF-NAS-MBV2-B, respectively. The latency measurement, the search and the evaluation hyperparameters are same with Appendix~\ref{more_details_of_experimental_settings}, except that the dropout rate of TF-NAS-MBV2-A and TF-NAS-MBV2-B are set to 0,2 and 0.1, respectively. The results are presented in Tab.\ref{tab:mbv2_results}. Compared with MobileNetV2~\cite{DBLP:conf/cvpr/SandlerHZZC18}, our TF-NAS-MBV2-A and TF-NAS-MBV2-B exceed their competitors by 0.6\% and 1.7\% on the top-1 accuracy with less latency. Moreover, under the same GPU latency, TF-NAS-MBV2-B outperforms MobileNetv3 0.75$\times$~\cite{DBLP:conf/iccv/HowardSCCCTWZPVLA19} by 0.4\% top-1 accuracy. These observations indicate the universality of TF-NAS to other search space. \vspace{-0.3cm}
\begin{table}
\begin{minipage}{0.48\linewidth}
\centering
\resizebox{1.0\textwidth}{!}{
\begin{tabular}{c|c|c|c|c|c} \hline
  Stage & Input & Operation & $C_{out}$ & Act & L \\
  \hline
  1  & $224^2\times3$  & $3\times3$ Conv & 32   & ReLU6 & 1 \\
  2  & $112^3\times32$ & MBInvRes        & 16   & ReLU6 & 1 \\
  3  & $112^2\times16$ & OPS             & 24   & ReLU6 & $[1, 2]$ \\
  4  & $56^2\times24$  & OPS             & 32   & ReLU6 & $[1, 3]$ \\
  5  & $28^2\times32$  & OPS             & 64   & ReLU6 & $[1, 4]$ \\
  6  & $14^2\times64$  & OPS             & 96   & ReLU6 & $[1, 4]$ \\
  7  & $14^2\times96$  & OPS             & 160  & ReLU6 & $[1, 4]$ \\
  8  & $7^2\times160$  & OPS             & 320  & ReLU6 & 1 \\
  9  & $7^2\times320$  & $1\times1$ Conv & 1280 & ReLU6 & 1 \\
  10 & $7^2\times1280$ & AvgPool         & 1280 & -     & 1 \\
  11 & $1280$          & Fc              & 1000 & -     & 1 \\
  \hline
  \end{tabular}}
\end{minipage}
\begin{minipage}{0.48\linewidth}
\centering
\resizebox{0.6\textwidth}{!}{
\begin{tabular}{l|c|c} \hline
  OPS & Kernel & Expansion \\
  \hline
  $k3\_e3$          & 3 & $[2, 4]$ \\
  $k5\_e3$          & 5 & $[2, 4]$ \\
  $k3\_e6$          & 3 & $[4, 8]$ \\
  $k5\_e6$          & 5 & $[4, 8]$ \\
  \hline
  \end{tabular}}
\end{minipage} \vspace{0.2cm}
\caption{\textbf{Left}: Macro architecture of the MobileNetV2-based supernet. ``OPS" denotes the operations to be searched. ``MBInvRes" is the basic block in~\cite{DBLP:conf/cvpr/SandlerHZZC18}. ``$C_{out}$" means the output channels. ``Act" denotes the activation function used in a stage. ``L" is the number of layers in a stage, where $[a, b]$ is a discrete interval. If necessary, the down-sampling occurs at the first operation of a stage. \textbf{Right}: Candidate operations to be searched. ``Expansion" defines the width of an operation and $[a, b]$ is a continuous interval.} \vspace{-1.4cm}
\label{tab:mbv2_macro_arch_and_ops}
\end{table}

\begin{table}
  \centering
  \resizebox{0.8\textwidth}{!}{
  \begin{tabular}{l|c|c|c|c} \hline
  Architecture & \tabincell{c}{Top-1\\Acc(\%)} & \tabincell{c}{GPU\\Latency} & \tabincell{c}{FLOPs\\(M)} & \tabincell{c}{Search Time\\(GPU days)} \\
  \hline
  MobileNetV2 1.4$\times$~\cite{DBLP:conf/cvpr/SandlerHZZC18}        & 74.7 & 16.18ms & 585  & -   \\
  TF-NAS-MBV2-A (Ours)                                               & 75.3 & 14.93ms & 445  & 1   \\
  \hline
  MobileNetV1~\cite{DBLP:journals/corr/HowardZCKWWAA17}              & 70.6 & 9.73ms  & 569 & -    \\
  MobileNetV2~\cite{DBLP:conf/cvpr/SandlerHZZC18}                    & 72.0 & 11.15ms & 300 & -    \\
  MobileNetV3 0.75$\times$~\cite{DBLP:conf/iccv/HowardSCCCTWZPVLA19} & 73.3 & 10.01ms & 155 & -    \\
  TF-NAS-MBV2-B (Ours)                                               & 73.7 & 10.06ms & 297 & 1    \\
  \hline
  \end{tabular}}\vspace{0.2cm}
  \caption{Results on MobileNetV2-based search space. For the GPU latency, we measure it with a batch size of 32 on a Titan RTX GPU.} \vspace{-1.2cm}
  \label{tab:mbv2_results}
\end{table}

\section{Differences with Previous Works}
\vspace{-0.2cm}
We compare our TF-NAS with current differentiable NAS for macro search in Tab.~\ref{tab:comparison_dnas}. Both TF-NAS and DenseNAS have three search freedoms, but they have three differences: 1) For the operation-level search, DenseNAS samples one path with the maximum architecture distribution parameter, increasing the risk of operation collapse. Our TF-NAS employ a bi-sampling algorithm to moderate the operation collapse. 2) DenseNAS couples the width-level and depth-level search together, where searching for depth is equivalent to searching for the layers with different widths. Our TF-NAS searches for depth in a sink-connecting space, which is independent of the width-level search, increasing the search flexibility. 3) DenseNAS adds additional layers and assembles them by dense connection to search for width. The former greatly increases the GPU memory and the search time. The later connects a layer to the following four layers with different widths, which means there are only four choices of width can be searched in each layer. Differently, our TF-NAS adaptively shrinks and expands the operation channels to control the architecture width, which has more width choices than DenseNAS and can search latency-satisfied architectures. Furthermore, as mentioned in Appendix~\ref{implementation_details_of_elasticity-scaling}, our approach does not increase additional GPU memory.


On the other hand, our elasticity-scaling strategy is inspired by MorphNet~\cite{DBLP:conf/cvpr/GordonENCWYC18}. The differences between them are as follows: 1) Our approach shrinks and expands a model in a progressively fine-gained manner, but MorphNet only employs a global manner. 2) Both shrinking and expanding in our approach are based on the channel importance, while MorphNet uses sparse regularizations in shrinking and a direct width multiplier in expanding. 3) Model weights are shared and can be reused in our approach, but the morphed model needs to be trained from scratch for the next morphing in MorphNet. \vspace{-0.2cm}

\begin{table*}
  \centering
  \resizebox{0.9\textwidth}{!}{
  \begin{tabular}{l|c|c|c|c|c} \hline
  Method & \tabincell{c}{Operation-level\\Search} & \tabincell{c}{Depth-level\\Search} & \tabincell{c}{Width-level\\Search} & \tabincell{c}{Search Time\\(GPU days)} & Searched on \\
  \hline
  RCNet~\cite{DBLP:conf/iccv/XiongMS19}             & $\surd$ & $\times$ & $\times$ & 8     & ImageNet \\
  FPNASNet~\cite{DBLP:conf/iccv/CuiCLLSJ19}         & $\surd$ & $\times$ & $\times$ & 0.83  & CIFAR10  \\
  ProxylessNAS~\cite{DBLP:conf/iclr/CaiZH19}        & $\surd$ & $\surd$  & $\times$ & 8.3   & ImageNet \\
  FBNet~\cite{DBLP:conf/cvpr/WuDZWSWTVJK19}         & $\surd$ & $\surd$  & $\times$ & 9     & ImageNet \\
  AtomNAS-A~\cite{DBLP:conf/iclr/MeiLLJYYY20}       & $\surd$ & $\times$ & $\surd$  & -     & ImageNet \\
  DenseNAS~\cite{DBLP:conf/cvpr/FangSZLLW20}        & $\surd$ & $\surd$  & $\surd$  & 3.8   & ImageNet \\
  TF-NAS (Ours)                                     & $\surd$ & $\surd$  & $\surd$  & 1.8   & ImageNet \\
  \hline
  \end{tabular}}\vspace{0.2cm}
  \caption{Comparisons with current differentiable NAS for macro search. The dataset searched on is ImageNet~\cite{DBLP:conf/cvpr/DengDSLL009} or CIFAR10~\cite{DBLP:journals/tr/KrizhevskyH09}.} \vspace{-1.2cm}
  \label{tab:comparison_dnas}
\end{table*}

\section{Sensitivity Analysis of $\lambda$}
\begin{figure}
  \centering
  \includegraphics[width=1.0\linewidth]{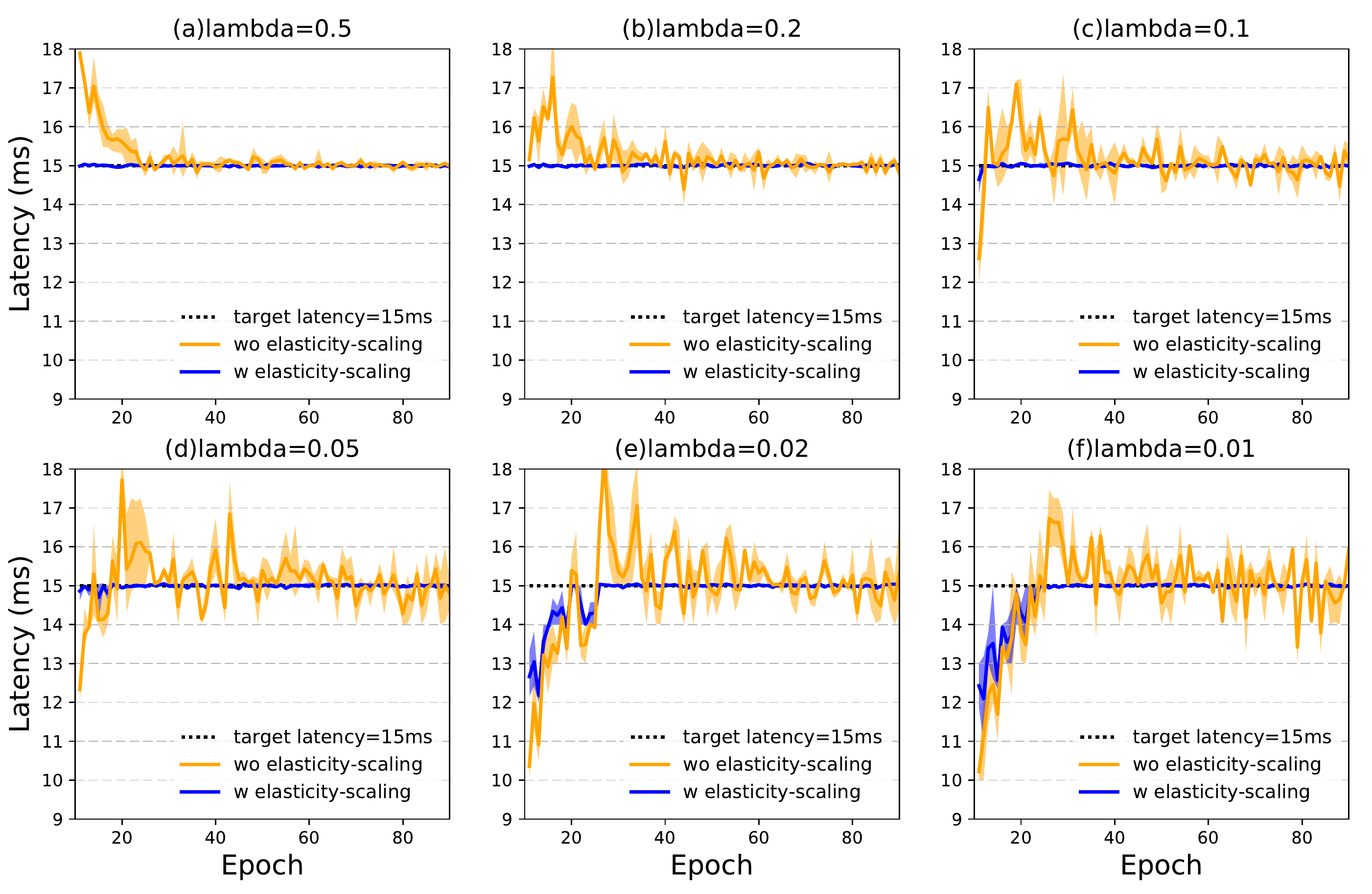} \vspace{-0.8cm}
  \caption{Searched latencies of various $\lambda$. All the search procedures are repeated 5 times, and we plot the mean, the maximum and the minimum. Zoom in for better view.} \vspace{-1.0cm}
  \label{fig:supp_diff_lam}
\end{figure}

There is a trade-off parameter $\lambda$ in our TF-NAS to balance between the accuracy and the latency. In this section, we analysis the sensitivity of $\lambda$ with or without our proposed elasticity-scaling strategy. Restricted to 15ms target latency, we set $\lambda$ to 0.5, 0.2, 0.1, 0.05, 0.02 and 0.01, respectively. As shown in Fig.~\ref{fig:supp_diff_lam}, without elasticity-scaling, $\lambda$ has a great impact on the latency of the searched architecture. On the one hand, small $\lambda$ (0.05, 0.02 and 0.01) hardly makes the searched architecture satisfy the target latency, where large jitters can be observed in Fig.~\ref{fig:supp_diff_lam}(d)-(f) (orange lines). On the other hand, large $\lambda$ (0.5, 0.2 and 0.1) makes the searched architecture slightly fluctuate down and up around the target 15ms after about 35 epochs (orange lines in Fig.~\ref{fig:supp_diff_lam}(a)-(c)), but cannot achieve precise target latency. By employing our elasticity-scaling strategy, all the settings of $\lambda$ can search architectures with perfect latency satisfaction (blue lines in Fig.~\ref{fig:supp_diff_lam}(a)-(f)). Although the target latency can be satisfied by elasticity-scaling, the accuracies of the searched architectures vary greatly under different $\lambda$. As shown in Tab.~\ref{tab:diff_lam}, $\lambda=0.1$ achieves the best top-1 accuracy. Thus, we set $\lambda$ to 0.1 for all the experiments. \vspace{-0.4cm}

\begin{table}
  \centering
  \begin{tabular}{l|c|c|c} \hline
  $\lambda$  & Top-1 Acc(\%) & GPU Latency & FLOPs(M) \\
  \hline
  0.5  & 76.0 & 15.01ms & 363 \\
  0.2  & 76.1 & 15.14ms & 372 \\
  0.1  & 76.3 & 15.06ms & 361 \\
  0.05 & 76.2 & 15.09ms & 361 \\
  0.02 & 75.9 & 15.10ms & 344 \\
  0.01 & 75.7 & 15.08ms & 366 \\
  \hline
  \end{tabular}\vspace{0.2cm}
  \caption{Comparisons with different trade-off $\lambda$.} \vspace{-1.6cm}
  \label{tab:diff_lam}
\end{table}

\section{Details of Searched Architectures}
The architecture details of our searched TF-NAS-A, TF-NAS-B, TF-NAS-C and TF-NAS-D are depicted in Tab.~\ref{tab:TF-NAS-A}, Tab.~\ref{tab:TF-NAS-B}, Tab.~\ref{tab:TF-NAS-C} and Tab.~\ref{tab:TF-NAS-D}, respectively. For architectures without SE module, i.e. TF-NAS-A-wose, TF-NAS-B-wose, TF-NAS-C-wose and TF-NAS-D-wose, the details are listed in Tab.~\ref{tab:TF-NAS-A-wose}, Tab.~\ref{tab:TF-NAS-B-wose}, Tab.~\ref{tab:TF-NAS-C-wose} and Tab.~\ref{tab:TF-NAS-D-wose}, respectively. Besides, Tab.~\ref{tab:TF-NAS-CPU-A} and Tab.~\ref{tab:TF-NAS-CPU-B} present the architectures of CPU constrained TF-NAS, i.e. TF-NAS-CPU-A and TF-NAS-CPU-B. Finally, TF-NAS-MBV2-A and TF-NAS-MBV2-B are summarized in Tab.~\ref{tab:TF-NAS-MBV2-A} and Tab.~\ref{tab:TF-NAS-MBV2-B}, respectively.

\begin{table*}
  \centering
  \resizebox{0.74\textwidth}{!}{
  \begin{tabular}{c|c|c|c|c|c|c|c} \hline
  Input & Operation & $C_{in}$ & $e \times C_{in}$ & $e_{se} \times C_{in}$ & $C_{out}$ & Act & Stride \\
  \hline
  $224^2\times3$   & $3\times3$ Conv & 3    & -    & -   & 32   & ReLU  & 2 \\
  $112^3\times32$  & MBInvRes\_$k3$  & 32   & 32   & 8   & 16   & ReLU  & 1 \\
  $112^2\times16$  & MBInvRes\_$k3$  & 16   & 83   & 32  & 24   & ReLU  & 2 \\
  $56^2\times24$   & MBInvRes\_$k5$  & 24   & 128  & 0   & 24   & ReLU  & 1 \\
  $56^2\times24$   & MBInvRes\_$k3$  & 24   & 138  & 48  & 40   & Swish & 2 \\
  $28^2\times40$   & MBInvRes\_$k3$  & 40   & 297  & 0   & 40   & Swish & 1 \\
  $28^2\times40$   & MBInvRes\_$k5$  & 40   & 170  & 80  & 40   & Swish & 1 \\
  $28^2\times40$   & MBInvRes\_$k5$  & 40   & 248  & 80  & 80   & Swish & 2 \\
  $14^2\times80$   & MBInvRes\_$k3$  & 80   & 500  & 0   & 80   & Swish & 1 \\
  $14^2\times80$   & MBInvRes\_$k3$  & 80   & 424  & 0   & 80   & Swish & 1 \\
  $14^2\times80$   & MBInvRes\_$k3$  & 80   & 477  & 0   & 80   & Swish & 1 \\
  $14^2\times80$   & MBInvRes\_$k3$  & 80   & 504  & 160 & 112  & Swish & 1 \\
  $14^2\times112$  & MBInvRes\_$k3$  & 112  & 796  & 0   & 112  & Swish & 1 \\
  $14^2\times112$  & MBInvRes\_$k3$  & 112  & 723  & 224 & 112  & Swish & 1 \\
  $14^2\times112$  & MBInvRes\_$k3$  & 112  & 555  & 224 & 112  & Swish & 1 \\
  $14^2\times112$  & MBInvRes\_$k3$  & 112  & 813  & 0   & 192  & Swish & 2 \\
  $7^2\times192$   & MBInvRes\_$k3$  & 192  & 1370 & 0   & 192  & Swish & 1 \\
  $7^2\times192$   & MBInvRes\_$k3$  & 192  & 1138 & 384 & 192  & Swish & 1 \\
  $7^2\times192$   & MBInvRes\_$k3$  & 192  & 1359 & 384 & 192  & Swish & 1 \\
  $7^2\times192$   & MBInvRes\_$k5$  & 192  & 1203 & 384 & 320  & Swish & 1 \\
  $7^2\times320$   & $1\times1$ Conv & 320  & -    & -   & 1280 & Swish & 1 \\
  $7^2\times1280$  & AvgPool         & 1280 & -    & -   & 1280 & -     & - \\
  $1280$           & Fc              & 1280 & -    & -   & 1000 & -     & - \\
  \hline
  \end{tabular}} \vspace{0.1cm}
  \caption{Architecture details of TF-NAS-A.}\vspace{-0.4cm}
  \label{tab:TF-NAS-A}
\end{table*}

\begin{table*}
  \centering
  \resizebox{0.74\textwidth}{!}{
  \begin{tabular}{c|c|c|c|c|c|c|c} \hline
  Input & Operation & $C_{in}$ & $e \times C_{in}$ & $e_{se} \times C_{in}$ & $C_{out}$ & Act & Stride \\
  \hline
  $224^2\times3$   & $3\times3$ Conv & 3    & -    & -   & 32   & ReLU  & 2 \\
  $112^3\times32$  & MBInvRes\_$k3$  & 32   & 32   & 8   & 16   & ReLU  & 1 \\
  $112^2\times16$  & MBInvRes\_$k3$  & 16   & 64   & 32  & 24   & ReLU  & 2 \\
  $56^2\times24$   & MBInvRes\_$k3$  & 24   & 118  & 48  & 24   & ReLU  & 1 \\
  $56^2\times24$   & MBInvRes\_$k5$  & 24   & 96   & 48  & 40   & Swish & 2 \\
  $28^2\times40$   & MBInvRes\_$k3$  & 40   & 203  & 80  & 40   & Swish & 1 \\
  $28^2\times40$   & MBInvRes\_$k5$  & 40   & 161  & 80  & 40   & Swish & 1 \\
  $28^2\times40$   & MBInvRes\_$k3$  & 40   & 224  & 0   & 80   & Swish & 2 \\
  $14^2\times80$   & MBInvRes\_$k5$  & 80   & 361  & 160 & 80   & Swish & 1 \\
  $14^2\times80$   & MBInvRes\_$k3$  & 80   & 323  & 160 & 80   & Swish & 1 \\
  $14^2\times80$   & MBInvRes\_$k3$  & 80   & 320  & 160 & 80   & Swish & 1 \\
  $14^2\times80$   & MBInvRes\_$k3$  & 80   & 324  & 160 & 112  & Swish & 1 \\
  $14^2\times112$  & MBInvRes\_$k3$  & 112  & 581  & 0   & 112  & Swish & 1 \\
  $14^2\times112$  & MBInvRes\_$k3$  & 112  & 482  & 224 & 112  & Swish & 1 \\
  $14^2\times112$  & MBInvRes\_$k3$  & 112  & 667  & 0   & 112  & Swish & 1 \\
  $14^2\times112$  & MBInvRes\_$k3$  & 112  & 579  & 0   & 192  & Swish & 2 \\
  $7^2\times192$   & MBInvRes\_$k3$  & 192  & 738  & 0   & 192  & Swish & 1 \\
  $7^2\times192$   & MBInvRes\_$k3$  & 192  & 1028 & 384 & 192  & Swish & 1 \\
  $7^2\times192$   & MBInvRes\_$k3$  & 192  & 1161 & 384 & 192  & Swish & 1 \\
  $7^2\times192$   & MBInvRes\_$k5$  & 192  & 881  & 384 & 320  & Swish & 1 \\
  $7^2\times320$   & $1\times1$ Conv & 320  & -    & -   & 1280 & Swish & 1 \\
  $7^2\times1280$  & AvgPool         & 1280 & -    & -   & 1280 & -     & - \\
  $1280$           & Fc              & 1280 & -    & -   & 1000 & -     & - \\
  \hline
  \end{tabular}}  \vspace{0.1cm}
  \caption{Architecture details of TF-NAS-B.}
  \label{tab:TF-NAS-B}
\end{table*}

\begin{table*}
  \centering
  \resizebox{0.74\textwidth}{!}{
  \begin{tabular}{c|c|c|c|c|c|c|c} \hline
  Input & Operation & $C_{in}$ & $e \times C_{in}$ & $e_{se} \times C_{in}$ & $C_{out}$ & Act & Stride \\
  \hline
  $224^2\times3$   & $3\times3$ Conv & 3    & -    & -   & 32   & ReLU  & 2 \\
  $112^3\times32$  & MBInvRes\_$k3$  & 32   & 32   & 8   & 16   & ReLU  & 1 \\
  $112^2\times16$  & MBInvRes\_$k5$  & 16   & 64   & 32  & 24   & ReLU  & 2 \\
  $56^2\times24$   & MBInvRes\_$k5$  & 24   & 48   & 24  & 40   & Swish & 2 \\
  $28^2\times40$   & MBInvRes\_$k5$  & 40   & 160  & 80  & 40   & Swish & 1 \\
  $28^2\times40$   & MBInvRes\_$k5$  & 40   & 160  & 80  & 40   & Swish & 1 \\
  $28^2\times40$   & MBInvRes\_$k5$  & 40   & 160  & 80  & 80   & Swish & 2 \\
  $14^2\times80$   & MBInvRes\_$k5$  & 80   & 320  & 0   & 80   & Swish & 1 \\
  $14^2\times80$   & MBInvRes\_$k5$  & 80   & 160  & 0   & 80   & Swish & 1 \\
  $14^2\times80$   & MBInvRes\_$k3$  & 80   & 320  & 160 & 80   & Swish & 1 \\
  $14^2\times80$   & MBInvRes\_$k3$  & 80   & 320  & 0   & 112  & Swish & 1 \\
  $14^2\times112$  & MBInvRes\_$k5$  & 112  & 448  & 224 & 112  & Swish & 1 \\
  $14^2\times112$  & MBInvRes\_$k3$  & 112  & 448  & 0   & 112  & Swish & 1 \\
  $14^2\times112$  & MBInvRes\_$k3$  & 112  & 448  & 224 & 112  & Swish & 1 \\
  $14^2\times112$  & MBInvRes\_$k3$  & 112  & 448  & 224 & 192  & Swish & 2 \\
  $7^2\times192$   & MBInvRes\_$k5$  & 192  & 768  & 384 & 192  & Swish & 1 \\
  $7^2\times192$   & MBInvRes\_$k5$  & 192  & 768  & 384 & 192  & Swish & 1 \\
  $7^2\times192$   & MBInvRes\_$k3$  & 192  & 384  & 192 & 192  & Swish & 1 \\
  $7^2\times192$   & MBInvRes\_$k5$  & 192  & 768  & 384 & 320  & Swish & 1 \\
  $7^2\times320$   & $1\times1$ Conv & 320  & -    & -   & 1280 & Swish & 1 \\
  $7^2\times1280$  & AvgPool         & 1280 & -    & -   & 1280 & -     & - \\
  $1280$           & Fc              & 1280 & -    & -   & 1000 & -     & - \\
  \hline
  \end{tabular}} \vspace{0.1cm}
  \caption{Architecture details of TF-NAS-C.}
  \label{tab:TF-NAS-C}
\end{table*}

\begin{table*}
  \centering
  \resizebox{0.74\textwidth}{!}{
  \begin{tabular}{c|c|c|c|c|c|c|c} \hline
  Input & Operation & $C_{in}$ & $e \times C_{in}$ & $e_{se} \times C_{in}$ & $C_{out}$ & Act & Stride \\
  \hline
  $224^2\times3$   & $3\times3$ Conv & 3    & -    & -   & 32   & ReLU  & 2 \\
  $112^3\times32$  & MBInvRes\_$k3$  & 32   & 32   & 8   & 16   & ReLU  & 1 \\
  $112^2\times16$  & MBInvRes\_$k3$  & 16   & 65   & 32  & 24   & ReLU  & 2 \\
  $56^2\times24$   & MBInvRes\_$k3$  & 24   & 63   & 0   & 24   & ReLU  & 1 \\
  $56^2\times24$   & MBInvRes\_$k3$  & 24   & 58   & 24  & 40   & Swish & 2 \\
  $28^2\times40$   & MBInvRes\_$k5$  & 40   & 106  & 0   & 40   & Swish & 1 \\
  $28^2\times40$   & MBInvRes\_$k5$  & 40   & 80   & 0   & 40   & Swish & 1 \\
  $28^2\times40$   & MBInvRes\_$k3$  & 40   & 192  & 80  & 80   & Swish & 2 \\
  $14^2\times80$   & MBInvRes\_$k3$  & 80   & 219  & 0   & 80   & Swish & 1 \\
  $14^2\times80$   & MBInvRes\_$k5$  & 80   & 320  & 0   & 80   & Swish & 1 \\
  $14^2\times80$   & MBInvRes\_$k3$  & 80   & 212  & 80  & 80   & Swish & 1 \\
  $14^2\times80$   & MBInvRes\_$k3$  & 80   & 165  & 0   & 112  & Swish & 1 \\
  $14^2\times112$  & MBInvRes\_$k5$  & 112  & 245  & 112 & 112  & Swish & 1 \\
  $14^2\times112$  & MBInvRes\_$k3$  & 112  & 292  & 112 & 112  & Swish & 1 \\
  $14^2\times112$  & MBInvRes\_$k3$  & 112  & 408  & 112 & 112  & Swish & 1 \\
  $14^2\times112$  & MBInvRes\_$k3$  & 112  & 538  & 0   & 192  & Swish & 2 \\
  $7^2\times192$   & MBInvRes\_$k5$  & 192  & 768  & 192 & 320  & Swish & 1 \\
  $7^2\times320$   & $1\times1$ Conv & 320  & -    & -   & 1280 & Swish & 1 \\
  $7^2\times1280$  & AvgPool         & 1280 & -    & -   & 1280 & -     & - \\
  $1280$           & Fc              & 1280 & -    & -   & 1000 & -     & - \\
  \hline
  \end{tabular}} \vspace{0.1cm}
  \caption{Architecture details of TF-NAS-D.}
  \label{tab:TF-NAS-D}
\end{table*}

\begin{table*}
  \centering
  \resizebox{0.74\textwidth}{!}{
  \begin{tabular}{c|c|c|c|c|c|c|c} \hline
  Input & Operation & $C_{in}$ & $e \times C_{in}$ & $e_{se} \times C_{in}$ & $C_{out}$ & Act & Stride \\
  \hline
  $224^2\times3$   & $3\times3$ Conv & 3    & -    & -   & 32   & ReLU  & 2 \\
  $112^3\times32$  & MBInvRes\_$k3$  & 32   & 32   & 0   & 16   & ReLU  & 1 \\
  $112^2\times16$  & MBInvRes\_$k3$  & 16   & 74   & 0   & 24   & ReLU  & 2 \\
  $56^2\times24$   & MBInvRes\_$k3$  & 24   & 127  & 0   & 24   & ReLU  & 1 \\
  $56^2\times24$   & MBInvRes\_$k3$  & 24   & 154  & 0   & 40   & Swish & 2 \\
  $28^2\times40$   & MBInvRes\_$k5$  & 40   & 239  & 0   & 40   & Swish & 1 \\
  $28^2\times40$   & MBInvRes\_$k5$  & 40   & 234  & 0   & 40   & Swish & 1 \\
  $28^2\times40$   & MBInvRes\_$k5$  & 40   & 270  & 0   & 80   & Swish & 2 \\
  $14^2\times80$   & MBInvRes\_$k3$  & 80   & 595  & 0   & 80   & Swish & 1 \\
  $14^2\times80$   & MBInvRes\_$k5$  & 80   & 506  & 0   & 80   & Swish & 1 \\
  $14^2\times80$   & MBInvRes\_$k3$  & 80   & 572  & 0   & 80   & Swish & 1 \\
  $14^2\times80$   & MBInvRes\_$k3$  & 80   & 640  & 0   & 112  & Swish & 1 \\
  $14^2\times112$  & MBInvRes\_$k3$  & 112  & 895  & 0   & 112  & Swish & 1 \\
  $14^2\times112$  & MBInvRes\_$k5$  & 112  & 802  & 0   & 112  & Swish & 1 \\
  $14^2\times112$  & MBInvRes\_$k3$  & 112  & 895  & 0   & 112  & Swish & 1 \\
  $14^2\times112$  & MBInvRes\_$k3$  & 112  & 817  & 0   & 192  & Swish & 2 \\
  $7^2\times192$   & MBInvRes\_$k5$  & 192  & 1536 & 0   & 192  & Swish & 1 \\
  $7^2\times192$   & MBInvRes\_$k3$  & 192  & 1281 & 0   & 192  & Swish & 1 \\
  $7^2\times192$   & MBInvRes\_$k5$  & 192  & 1495 & 0   & 192  & Swish & 1 \\
  $7^2\times192$   & MBInvRes\_$k5$  & 192  & 1536 & 0   & 320  & Swish & 1 \\
  $7^2\times320$   & $1\times1$ Conv & 320  & -    & -   & 1280 & Swish & 1 \\
  $7^2\times1280$  & AvgPool         & 1280 & -    & -   & 1280 & -     & - \\
  $1280$           & Fc              & 1280 & -    & -   & 1000 & -     & - \\
  \hline
  \end{tabular}} \vspace{0.1cm}
  \caption{Architecture details of TF-NAS-A-wose.}
  \label{tab:TF-NAS-A-wose}
\end{table*}

\begin{table*}
  \centering
  \resizebox{0.74\textwidth}{!}{
  \begin{tabular}{c|c|c|c|c|c|c|c} \hline
  Input & Operation & $C_{in}$ & $e \times C_{in}$ & $e_{se} \times C_{in}$ & $C_{out}$ & Act & Stride \\
  \hline
  $224^2\times3$   & $3\times3$ Conv & 3    & -    & -   & 32   & ReLU  & 2 \\
  $112^3\times32$  & MBInvRes\_$k3$  & 32   & 32   & 0   & 16   & ReLU  & 1 \\
  $112^2\times16$  & MBInvRes\_$k5$  & 16   & 65   & 0   & 24   & ReLU  & 2 \\
  $56^2\times24$   & MBInvRes\_$k5$  & 24   & 98   & 0   & 24   & ReLU  & 1 \\
  $56^2\times24$   & MBInvRes\_$k5$  & 24   & 104  & 0   & 40   & Swish & 2 \\
  $28^2\times40$   & MBInvRes\_$k5$  & 40   & 136  & 0   & 40   & Swish & 1 \\
  $28^2\times40$   & MBInvRes\_$k5$  & 40   & 135  & 0   & 40   & Swish & 1 \\
  $28^2\times40$   & MBInvRes\_$k3$  & 40   & 248  & 0   & 80   & Swish & 2 \\
  $14^2\times80$   & MBInvRes\_$k3$  & 80   & 409  & 0   & 80   & Swish & 1 \\
  $14^2\times80$   & MBInvRes\_$k3$  & 80   & 530  & 0   & 80   & Swish & 1 \\
  $14^2\times80$   & MBInvRes\_$k5$  & 80   & 251  & 0   & 80   & Swish & 1 \\
  $14^2\times80$   & MBInvRes\_$k3$  & 80   & 498  & 0   & 112  & Swish & 1 \\
  $14^2\times112$  & MBInvRes\_$k3$  & 112  & 639  & 0   & 112  & Swish & 1 \\
  $14^2\times112$  & MBInvRes\_$k5$  & 112  & 573  & 0   & 112  & Swish & 1 \\
  $14^2\times112$  & MBInvRes\_$k3$  & 112  & 718  & 0   & 112  & Swish & 1 \\
  $14^2\times112$  & MBInvRes\_$k5$  & 112  & 896  & 0   & 192  & Swish & 2 \\
  $7^2\times192$   & MBInvRes\_$k5$  & 192  & 1209 & 0   & 192  & Swish & 1 \\
  $7^2\times192$   & MBInvRes\_$k3$  & 192  & 1276 & 0   & 192  & Swish & 1 \\
  $7^2\times192$   & MBInvRes\_$k3$  & 192  & 1536 & 0   & 192  & Swish & 1 \\
  $7^2\times192$   & MBInvRes\_$k5$  & 192  & 1526 & 0   & 320  & Swish & 1 \\
  $7^2\times320$   & $1\times1$ Conv & 320  & -    & -   & 1280 & Swish & 1 \\
  $7^2\times1280$  & AvgPool         & 1280 & -    & -   & 1280 & -     & - \\
  $1280$           & Fc              & 1280 & -    & -   & 1000 & -     & - \\
  \hline
  \end{tabular}} \vspace{0.1cm}
  \caption{Architecture details of TF-NAS-B-wose.}
  \label{tab:TF-NAS-B-wose}
\end{table*}

\begin{table*}
  \centering
  \resizebox{0.74\textwidth}{!}{
  \begin{tabular}{c|c|c|c|c|c|c|c} \hline
  Input & Operation & $C_{in}$ & $e \times C_{in}$ & $e_{se} \times C_{in}$ & $C_{out}$ & Act & Stride \\
  \hline
  $224^2\times3$   & $3\times3$ Conv & 3    & -    & -   & 32   & ReLU  & 2 \\
  $112^3\times32$  & MBInvRes\_$k3$  & 32   & 32   & 0   & 16   & ReLU  & 1 \\
  $112^2\times16$  & MBInvRes\_$k5$  & 16   & 64   & 0   & 24   & ReLU  & 2 \\
  $56^2\times24$   & MBInvRes\_$k5$  & 24   & 96   & 0   & 24   & ReLU  & 1 \\
  $56^2\times24$   & MBInvRes\_$k5$  & 24   & 48   & 0   & 40   & Swish & 2 \\
  $28^2\times40$   & MBInvRes\_$k5$  & 40   & 160  & 0   & 40   & Swish & 1 \\
  $28^2\times40$   & MBInvRes\_$k5$  & 40   & 160  & 0   & 40   & Swish & 1 \\
  $28^2\times40$   & MBInvRes\_$k5$  & 40   & 160  & 0   & 80   & Swish & 2 \\
  $14^2\times80$   & MBInvRes\_$k3$  & 80   & 320  & 0   & 80   & Swish & 1 \\
  $14^2\times80$   & MBInvRes\_$k5$  & 80   & 320  & 0   & 80   & Swish & 1 \\
  $14^2\times80$   & MBInvRes\_$k5$  & 80   & 320  & 0   & 80   & Swish & 1 \\
  $14^2\times80$   & MBInvRes\_$k3$  & 80   & 320  & 0   & 112  & Swish & 1 \\
  $14^2\times112$  & MBInvRes\_$k3$  & 112  & 448  & 0   & 112  & Swish & 1 \\
  $14^2\times112$  & MBInvRes\_$k3$  & 112  & 448  & 0   & 112  & Swish & 1 \\
  $14^2\times112$  & MBInvRes\_$k3$  & 112  & 448  & 0   & 112  & Swish & 1 \\
  $14^2\times112$  & MBInvRes\_$k3$  & 112  & 448  & 0   & 192  & Swish & 2 \\
  $7^2\times192$   & MBInvRes\_$k5$  & 192  & 768  & 0   & 192  & Swish & 1 \\
  $7^2\times192$   & MBInvRes\_$k3$  & 192  & 768  & 0   & 192  & Swish & 1 \\
  $7^2\times192$   & MBInvRes\_$k5$  & 192  & 768  & 0   & 192  & Swish & 1 \\
  $7^2\times192$   & MBInvRes\_$k5$  & 192  & 768  & 0   & 320  & Swish & 1 \\
  $7^2\times320$   & $1\times1$ Conv & 320  & -    & -   & 1280 & Swish & 1 \\
  $7^2\times1280$  & AvgPool         & 1280 & -    & -   & 1280 & -     & - \\
  $1280$           & Fc              & 1280 & -    & -   & 1000 & -     & - \\
  \hline
  \end{tabular}} \vspace{0.1cm}
  \caption{Architecture details of TF-NAS-C-wose.}
  \label{tab:TF-NAS-C-wose}
\end{table*}

\begin{table*}
  \centering
  \resizebox{0.74\textwidth}{!}{
  \begin{tabular}{c|c|c|c|c|c|c|c} \hline
  Input & Operation & $C_{in}$ & $e \times C_{in}$ & $e_{se} \times C_{in}$ & $C_{out}$ & Act & Stride \\
  \hline
  $224^2\times3$   & $3\times3$ Conv & 3    & -    & -   & 32   & ReLU  & 2 \\
  $112^3\times32$  & MBInvRes\_$k3$  & 32   & 32   & 0   & 16   & ReLU  & 1 \\
  $112^2\times16$  & MBInvRes\_$k5$  & 16   & 42   & 0   & 24   & ReLU  & 2 \\
  $56^2\times24$   & MBInvRes\_$k3$  & 24   & 48   & 0   & 24   & ReLU  & 1 \\
  $56^2\times24$   & MBInvRes\_$k5$  & 24   & 67   & 0   & 40   & Swish & 2 \\
  $28^2\times40$   & MBInvRes\_$k3$  & 40   & 117  & 0   & 40   & Swish & 1 \\
  $28^2\times40$   & MBInvRes\_$k5$  & 40   & 105  & 0   & 40   & Swish & 1 \\
  $28^2\times40$   & MBInvRes\_$k3$  & 40   & 104  & 0   & 80   & Swish & 2 \\
  $14^2\times80$   & MBInvRes\_$k3$  & 80   & 214  & 0   & 80   & Swish & 1 \\
  $14^2\times80$   & MBInvRes\_$k5$  & 80   & 194  & 0   & 80   & Swish & 1 \\
  $14^2\times80$   & MBInvRes\_$k3$  & 80   & 234  & 0   & 112  & Swish & 1 \\
  $14^2\times112$  & MBInvRes\_$k3$  & 112  & 228  & 0   & 112  & Swish & 1 \\
  $14^2\times112$  & MBInvRes\_$k3$  & 112  & 457  & 0   & 112  & Swish & 1 \\
  $14^2\times112$  & MBInvRes\_$k3$  & 112  & 457  & 0   & 112  & Swish & 1 \\
  $14^2\times112$  & MBInvRes\_$k3$  & 112  & 633  & 0   & 192  & Swish & 2 \\
  $7^2\times192$   & MBInvRes\_$k5$  & 192  & 973  & 0   & 192  & Swish & 1 \\
  $7^2\times192$   & MBInvRes\_$k5$  & 192  & 1081 & 0   & 192  & Swish & 1 \\
  $7^2\times192$   & MBInvRes\_$k5$  & 192  & 1116 & 0   & 192  & Swish & 1 \\
  $7^2\times192$   & MBInvRes\_$k5$  & 192  & 1161 & 0   & 320  & Swish & 1 \\
  $7^2\times320$   & $1\times1$ Conv & 320  & -    & -   & 1280 & Swish & 1 \\
  $7^2\times1280$  & AvgPool         & 1280 & -    & -   & 1280 & -     & - \\
  $1280$           & Fc              & 1280 & -    & -   & 1000 & -     & - \\
  \hline
  \end{tabular}} \vspace{0.1cm}
  \caption{Architecture details of TF-NAS-D-wose.}
  \label{tab:TF-NAS-D-wose}
\end{table*}

\begin{table*}
  \centering
  \resizebox{0.74\textwidth}{!}{
  \begin{tabular}{c|c|c|c|c|c|c|c} \hline
  Input & Operation & $C_{in}$ & $e \times C_{in}$ & $e_{se} \times C_{in}$ & $C_{out}$ & Act & Stride \\
  \hline
  $224^2\times3$   & $3\times3$ Conv & 3    & -    & -   & 32   & ReLU  & 2 \\
  $112^3\times32$  & MBInvRes\_$k3$  & 32   & 32   & 8   & 16   & ReLU  & 1 \\
  $112^2\times16$  & MBInvRes\_$k3$  & 16   & 64   & 32  & 24   & ReLU  & 2 \\
  $56^2\times24$   & MBInvRes\_$k5$  & 24   & 96   & 48  & 24   & ReLU  & 1 \\
  $56^2\times24$   & MBInvRes\_$k5$  & 24   & 96   & 48  & 40   & Swish & 2 \\
  $28^2\times40$   & MBInvRes\_$k5$  & 40   & 160  & 80  & 40   & Swish & 1 \\
  $28^2\times40$   & MBInvRes\_$k5$  & 40   & 160  & 80  & 40   & Swish & 1 \\
  $28^2\times40$   & MBInvRes\_$k5$  & 40   & 160  & 80  & 80   & Swish & 2 \\
  $14^2\times80$   & MBInvRes\_$k3$  & 80   & 320  & 160 & 80   & Swish & 1 \\
  $14^2\times80$   & MBInvRes\_$k3$  & 80   & 160  & 80  & 80   & Swish & 1 \\
  $14^2\times80$   & MBInvRes\_$k3$  & 80   & 320  & 160 & 80   & Swish & 1 \\
  $14^2\times80$   & MBInvRes\_$k3$  & 80   & 320  & 160 & 112  & Swish & 1 \\
  $14^2\times112$  & MBInvRes\_$k3$  & 112  & 448  & 224 & 112  & Swish & 1 \\
  $14^2\times112$  & MBInvRes\_$k5$  & 112  & 448  & 224 & 112  & Swish & 1 \\
  $14^2\times112$  & MBInvRes\_$k5$  & 112  & 224  & 0   & 112  & Swish & 1 \\
  $14^2\times112$  & MBInvRes\_$k3$  & 112  & 448  & 224 & 192  & Swish & 2 \\
  $7^2\times192$   & MBInvRes\_$k3$  & 192  & 768  & 384 & 192  & Swish & 1 \\
  $7^2\times192$   & MBInvRes\_$k3$  & 192  & 768  & 384 & 192  & Swish & 1 \\
  $7^2\times192$   & MBInvRes\_$k3$  & 192  & 768  & 384 & 192  & Swish & 1 \\
  $7^2\times192$   & MBInvRes\_$k5$  & 192  & 768  & 384 & 320  & Swish & 1 \\
  $7^2\times320$   & $1\times1$ Conv & 320  & -    & -   & 1280 & Swish & 1 \\
  $7^2\times1280$  & AvgPool         & 1280 & -    & -   & 1280 & -     & - \\
  $1280$           & Fc              & 1280 & -    & -   & 1000 & -     & - \\
  \hline
  \end{tabular}} \vspace{0.1cm}
  \caption{Architecture details of TF-NAS-CPU-A.}
  \label{tab:TF-NAS-CPU-A}
\end{table*}

\begin{table*}
  \centering
  \resizebox{0.74\textwidth}{!}{
  \begin{tabular}{c|c|c|c|c|c|c|c} \hline
  Input & Operation & $C_{in}$ & $e \times C_{in}$ & $e_{se} \times C_{in}$ & $C_{out}$ & Act & Stride \\
  \hline
  $224^2\times3$   & $3\times3$ Conv & 3    & -    & -   & 32   & ReLU  & 2 \\
  $112^3\times32$  & MBInvRes\_$k3$  & 32   & 32   & 8   & 16   & ReLU  & 1 \\
  $112^2\times16$  & MBInvRes\_$k3$  & 16   & 64   & 32  & 24   & ReLU  & 2 \\
  $56^2\times24$   & MBInvRes\_$k3$  & 24   & 96   & 0   & 24   & ReLU  & 1 \\
  $56^2\times24$   & MBInvRes\_$k3$  & 24   & 96   & 48  & 40   & Swish & 2 \\
  $28^2\times40$   & MBInvRes\_$k5$  & 40   & 80   & 0   & 40   & Swish & 1 \\
  $28^2\times40$   & MBInvRes\_$k3$  & 40   & 80   & 0   & 40   & Swish & 1 \\
  $28^2\times40$   & MBInvRes\_$k5$  & 40   & 160  & 0   & 80   & Swish & 2 \\
  $14^2\times80$   & MBInvRes\_$k3$  & 80   & 320  & 0   & 80   & Swish & 1 \\
  $14^2\times80$   & MBInvRes\_$k3$  & 80   & 160  & 80  & 80   & Swish & 1 \\
  $14^2\times80$   & MBInvRes\_$k5$  & 80   & 320  & 0   & 80   & Swish & 1 \\
  $14^2\times80$   & MBInvRes\_$k3$  & 80   & 320  & 0   & 112  & Swish & 1 \\
  $14^2\times112$  & MBInvRes\_$k3$  & 112  & 448  & 224 & 192  & Swish & 2 \\
  $7^2\times192$   & MBInvRes\_$k3$  & 192  & 768  & 384 & 192  & Swish & 1 \\
  $7^2\times192$   & MBInvRes\_$k5$  & 192  & 768  & 0   & 192  & Swish & 1 \\
  $7^2\times192$   & MBInvRes\_$k3$  & 192  & 768  & 0   & 192  & Swish & 1 \\
  $7^2\times192$   & MBInvRes\_$k3$  & 192  & 768  & 384 & 320  & Swish & 1 \\
  $7^2\times320$   & $1\times1$ Conv & 320  & -    & -   & 1280 & Swish & 1 \\
  $7^2\times1280$  & AvgPool         & 1280 & -    & -   & 1280 & -     & - \\
  $1280$           & Fc              & 1280 & -    & -   & 1000 & -     & - \\
  \hline
  \end{tabular}} \vspace{0.1cm}
  \caption{Architecture details of TF-NAS-CPU-B.}
  \label{tab:TF-NAS-CPU-B}
\end{table*}

\begin{table*}
  \centering
  \resizebox{0.74\textwidth}{!}{
  \begin{tabular}{c|c|c|c|c|c|c|c} \hline
  Input & Operation & $C_{in}$ & $e \times C_{in}$ & $e_{se} \times C_{in}$ & $C_{out}$ & Act & Stride \\
  \hline
  $224^2\times3$   & $3\times3$ Conv & 3    & -    & -   & 32   & ReLU6 & 2 \\
  $112^3\times32$  & MBInvRes\_$k3$  & 32   & 32   & 0   & 16   & ReLU6 & 1 \\
  $112^2\times16$  & MBInvRes\_$k5$  & 16   & 106  & 0   & 24   & ReLU6 & 2 \\
  $56^2\times24$   & MBInvRes\_$k3$  & 24   & 177  & 0   & 24   & ReLU6 & 1 \\
  $56^2\times24$   & MBInvRes\_$k5$  & 24   & 192  & 0   & 32   & ReLU6 & 2 \\
  $28^2\times32$   & MBInvRes\_$k5$  & 32   & 249  & 0   & 32   & ReLU6 & 1 \\
  $28^2\times32$   & MBInvRes\_$k3$  & 32   & 254  & 0   & 32   & ReLU6 & 1 \\
  $28^2\times32$   & MBInvRes\_$k3$  & 32   & 256  & 0   & 64   & ReLU6 & 2 \\
  $14^2\times64$   & MBInvRes\_$k3$  & 64   & 512  & 0   & 64   & ReLU6 & 1 \\
  $14^2\times64$   & MBInvRes\_$k5$  & 64   & 512  & 0   & 64   & ReLU6 & 1 \\
  $14^2\times64$   & MBInvRes\_$k3$  & 64   & 512  & 0   & 64   & ReLU6 & 1 \\
  $14^2\times64$   & MBInvRes\_$k3$  & 64   & 512  & 0   & 96   & ReLU6 & 1 \\
  $14^2\times96$   & MBInvRes\_$k5$  & 96   & 768  & 0   & 96   & ReLU6 & 1 \\
  $14^2\times96$   & MBInvRes\_$k3$  & 96   & 768  & 0   & 96   & ReLU6 & 1 \\
  $14^2\times96$   & MBInvRes\_$k3$  & 96   & 768  & 0   & 96   & ReLU6 & 1 \\
  $14^2\times96$   & MBInvRes\_$k3$  & 96   & 768  & 0   & 160  & ReLU6 & 2 \\
  $7^2\times160$   & MBInvRes\_$k5$  & 160  & 1280 & 0   & 160  & ReLU6 & 1 \\
  $7^2\times160$   & MBInvRes\_$k5$  & 160  & 1280 & 0   & 160  & ReLU6 & 1 \\
  $7^2\times160$   & MBInvRes\_$k3$  & 160  & 1280 & 0   & 160  & ReLU6 & 1 \\
  $7^2\times160$   & MBInvRes\_$k3$  & 160  & 1280 & 0   & 320  & ReLU6 & 1 \\
  $7^2\times320$   & $1\times1$ Conv & 320  & -    & -   & 1280 & ReLU6 & 1 \\
  $7^2\times1280$  & AvgPool         & 1280 & -    & -   & 1280 & -     & - \\
  $1280$           & Fc              & 1280 & -    & -   & 1000 & -     & - \\
  \hline
  \end{tabular}} \vspace{0.1cm}
  \caption{Architecture details of TF-NAS-MBV2-A.}
  \label{tab:TF-NAS-MBV2-A}
\end{table*}

\begin{table*}
  \centering
  \resizebox{0.74\textwidth}{!}{
  \begin{tabular}{c|c|c|c|c|c|c|c} \hline
  Input & Operation & $C_{in}$ & $e \times C_{in}$ & $e_{se} \times C_{in}$ & $C_{out}$ & Act & Stride \\
  \hline
  $224^2\times3$   & $3\times3$ Conv & 3    & -    & -   & 32   & ReLU6 & 2 \\
  $112^3\times32$  & MBInvRes\_$k3$  & 32   & 32   & 0   & 16   & ReLU6 & 1 \\
  $112^2\times16$  & MBInvRes\_$k5$  & 16   & 64   & 0   & 24   & ReLU6 & 2 \\
  $56^2\times24$   & MBInvRes\_$k5$  & 24   & 96   & 0   & 24   & ReLU6 & 1 \\
  $56^2\times24$   & MBInvRes\_$k5$  & 24   & 96   & 0   & 32   & ReLU6 & 2 \\
  $28^2\times32$   & MBInvRes\_$k3$  & 32   & 159  & 0   & 32   & ReLU6 & 1 \\
  $28^2\times32$   & MBInvRes\_$k5$  & 32   & 128  & 0   & 32   & ReLU6 & 1 \\
  $28^2\times32$   & MBInvRes\_$k5$  & 32   & 153  & 0   & 64   & ReLU6 & 2 \\
  $14^2\times64$   & MBInvRes\_$k3$  & 64   & 336  & 0   & 64   & ReLU6 & 1 \\
  $14^2\times64$   & MBInvRes\_$k5$  & 64   & 256  & 0   & 64   & ReLU6 & 1 \\
  $14^2\times64$   & MBInvRes\_$k5$  & 64   & 256  & 0   & 64   & ReLU6 & 1 \\
  $14^2\times64$   & MBInvRes\_$k5$  & 64   & 301  & 0   & 96   & ReLU6 & 1 \\
  $14^2\times96$   & MBInvRes\_$k3$  & 96   & 439  & 0   & 96   & ReLU6 & 1 \\
  $14^2\times96$   & MBInvRes\_$k3$  & 96   & 459  & 0   & 96   & ReLU6 & 1 \\
  $14^2\times96$   & MBInvRes\_$k5$  & 96   & 386  & 0   & 96   & ReLU6 & 1 \\
  $14^2\times96$   & MBInvRes\_$k3$  & 96   & 595  & 0   & 160  & ReLU6 & 2 \\
  $7^2\times160$   & MBInvRes\_$k5$  & 160  & 852  & 0   & 160  & ReLU6 & 1 \\
  $7^2\times160$   & MBInvRes\_$k3$  & 160  & 1004 & 0   & 160  & ReLU6 & 1 \\
  $7^2\times160$   & MBInvRes\_$k5$  & 160  & 1037 & 0   & 160  & ReLU6 & 1 \\
  $7^2\times160$   & MBInvRes\_$k5$  & 160  & 897  & 0   & 320  & ReLU6 & 1 \\
  $7^2\times320$   & $1\times1$ Conv & 320  & -    & -   & 1280 & ReLU6 & 1 \\
  $7^2\times1280$  & AvgPool         & 1280 & -    & -   & 1280 & -     & - \\
  $1280$           & Fc              & 1280 & -    & -   & 1000 & -     & - \\
  \hline
  \end{tabular}} \vspace{0.1cm}
  \caption{Architecture details of TF-NAS-MBV2-B.}
  \label{tab:TF-NAS-MBV2-B}
\end{table*}

\end{document}